\documentclass[11pt]{article}

\usepackage[final]{acl}

\usepackage{times}
\usepackage{latexsym}

\usepackage[T1]{fontenc}

\usepackage[utf8]{inputenc}

\usepackage{microtype}

\usepackage{inconsolata}

\usepackage{times}
\usepackage{latexsym}
\usepackage[T1]{fontenc}
\usepackage[utf8]{inputenc}
\usepackage{microtype}
\usepackage{inconsolata}
\usepackage{amsmath}
\usepackage{amssymb}
\usepackage{bbm}       
\usepackage{booktabs}
\usepackage{graphicx}
\usepackage[table]{xcolor}
\usepackage{colortbl}
\usepackage{enumitem}
\setlist[]{left=0mm,noitemsep,topsep=0mm}
\usepackage{float}
\usepackage{hyperref}
\usepackage{cleveref}
\usepackage{xspace}
\usepackage{multirow}
\usepackage{tcolorbox}
\tcbuselibrary{breakable, skins}
\usepackage{tikz}
\usepackage{kotex}       
\usepackage{fontawesome5}
\usepackage{subcaption}
\usepackage{CJKutf8}

\definecolor{MinorNoiseColor}{HTML}{B2A4FF}
\definecolor{tPurple}{HTML}{B2A4FF} % minor_noise

\definecolor{SourceCopyColor}{HTML}{FEE440}
\definecolor{tYellow}{HTML}{FEE440} % source_copy

\definecolor{WrongLangColor}{HTML}{71C792}
\definecolor{tGreen}{HTML}{71C792} % wrong_language

\definecolor{UnalignedColor}{HTML}{EBAF66}
\definecolor{tOrange}{HTML}{EBAF66} % unaligned_sent

\definecolor{HalluColor}{HTML}{70d6ff}
\definecolor{tBlue}{HTML}{70d6ff} % hallucination

\definecolor{SentReverseColor}{HTML}{FFC6FF}
\definecolor{tPink}{HTML}{FFC6FF} % wrong_sentiment

\definecolor{promptbg}{RGB}{245, 247, 250}
\definecolor{promptborder}{RGB}{100, 130, 180}
\definecolor{prompttitle}{RGB}{40, 60, 110}

\definecolor{shadeLow}{HTML}{FFF4E8}
\definecolor{shadeMid}{HTML}{FED8A8}
\definecolor{shadeHi}{HTML}{FA9C5B}
\definecolor{shadeMax}{HTML}{E65100}
\newcommand{\tikzlabel}[2]{%
    \tikz[baseline=(T.base)]\node(T)[
        fill=#1,
        rounded corners=2pt,
        inner sep=2pt,
        text=black,
        font=\bfseries,
    ]{#2};%
}

\tcbset{
  on line,
  boxsep=0pt, left=2pt, right=2pt, top=1pt, bottom=1pt,
  arc=1pt, boxrule=0pt,
}
\newcommand{\circled}[2]{%
    \tikz[baseline=(char.base)]{
        \node[shape=circle, draw, inner sep=1pt, 
              font=\small\bfseries, line width=0.6pt, fill=#1] 
        (char) {#2};
    }%
}

\newtcolorbox{promptbox}[1]{
    enhanced jigsaw, 
    colback=promptbg,
    colframe=promptborder,
    coltitle=yellow,
    colbacktitle=prompttitle,
    title=#1,
    fonttitle=\bfseries\small,
    fontupper=\small\ttfamily,
    boxrule=0.5pt,
    arc=2pt,
    left=8pt, right=8pt,
    top=6pt, bottom=6pt,
    attach boxed title to top center={yshift=0pt},
    boxed title style={
        colback=prompttitle,
        boxrule=0pt,
        arc=2pt,
    },
    title style={
        top=8pt, bottom=8pt,   
        left=12pt, right=12pt,  
    },
    before skip=8pt, after skip=8pt,
}

\newcommand{\xcomet}{x\textsc{comet-xxl}\xspace}            % xCOMET
 
\newcommand{\xcometkiwi}{\textsc{CometKiwi-xxl}\xspace}  
\newcommand{\metricx}{\textsc{metricX-24-xxl}\xspace}          % METRICX
\newcommand{\metricxqe}{\textsc{metricX-24-qe-xxl}\xspace}     % METRICX-QE
\newcommand{\chrf}{chr\textsc{F}\xspace}
\newcommand{\chrfpp}{chr\textsc{F++}\xspace}
\newcommand{\bertscore}{\textsc{BERTScore}\xspace}
\newcommand{\unite}{\textsc{UniTE}\xspace}
\newcommand{\uniteqe}{\textsc{UniTE-QE}\xspace}

\newcommand{\ter}{\textsc{ter}\xspace}
\newcommand{\mGpt}{\textsc{GPT-5.5}\xspace}
\newcommand{\mQwen}{\textsc{Qwen3.5-27B}\xspace}
\newcommand{\mirrorbench}{\textsc{ReBEal}\xspace}

\newcommand{\MinorNoiseCat}{\tikzlabel{tPink!80}{\texttt{\textcolor{black}{MINOR NOISE}}}\xspace}
\newcommand{\SourceCopyCat}{\tikzlabel{tPink!80}{\texttt{\textcolor{black}{SOURCE COPY}}}\xspace}
\newcommand{\WrongLangCat}{\tikzlabel{tGreen!70}{\texttt{\textcolor{black}{WRONG LANGUAGE}}}\xspace}
\newcommand{\UnalignedCat}{\tikzlabel{tGreen!70}{\texttt{\textcolor{black}{UNALIGNED SENT}}}\xspace}
\newcommand{\ContentAddCat}{\tikzlabel{tBlue!70}{\texttt{\textcolor{black}{CONTENT ADDITION}}}\xspace}
\newcommand{\PolarityCat}{\tikzlabel{tBlue!70}{\texttt{\textcolor{black}{POLARITY FLIP}}}\xspace}
\newcommand{\EntitySwapCat}{\tikzlabel{tBlue!70}{\texttt{\textcolor{black}{ENTITY SWAP}}}\xspace}
\newcommand{\ArgSwapCat}{\EntitySwapCat}
\newcommand{\HalluCat}{\ContentAddCat}
\newcommand{\Cat}[1]{\tikzlabel{gray!30}{\texttt{\textcolor{black}{\footnotesize #1}}}\xspace}

\newcommand{\sLow}[1]{\cellcolor{shadeLow}#1}
\newcommand{\sMid}[1]{\cellcolor{shadeMid}#1}
\newcommand{\sHi}[1]{\cellcolor{shadeHi}#1}
\newcommand{\sMax}[1]{\cellcolor{shadeMax}#1}

\newcommand{\hl}[1]{{\setlength{\fboxsep}{1pt}\colorbox{yellow!30}{#1}}}
\newcommand{\enlist}[1]{\circled{gray!20}{#1}\xspace}

\Crefname{appendix}{Appendix}{Appendices}

\title{\mirrorbench: Human References Are No Longer the Best Translation, \\and Metrics Do Not Notice}

\author{
  Ahrii Kim\textsuperscript{1} \quad
  Chanjun Park\textsuperscript{2 $\star$} \quad
  Seong-heum Kim\textsuperscript{1,3 $\star$} \\
  \textsuperscript{1}AI-Bio Convergence Research Inst. \quad
  \textsuperscript{2}School of Software \\
  \textsuperscript{3}Dept. of Intelligent Semiconductors \\
  Soongsil University \\
  \texttt{\{ahriikim,chanjun.park,seongheum\}@ssu.ac.kr}
}

\begin{document}
\maketitle

\begingroup
\renewcommand{\thefootnote}{$\star$}
\footnotetext{Corresponding authors.}
\endgroup

\begin{abstract}
Reference-based MT evaluation assumes the human reference is correct and better than the systems it scores. Both assumptions now fail, since the human reference ranks first in none of the 2024 WMT MQM campaigns and carries annotated errors in 19--87\% of its segments. We introduce \mirrorbench (\textbf{Re}ference \textbf{B}ias \textbf{Ev}\textbf{AL}uation), a benchmark that probes how MT metrics respond to seven types of controlled reference contamination, drawn from the error categories those campaigns record. Across eight widely used metrics and three language pairs, recent learned metrics (\textsc{xComet}, \textsc{MetricX}) prove more robust than their predecessors, but each family fails in its own way. Surface metrics show no length sensitivity, so contamination hides in longer segments, while learned metrics accept references in unrelated languages as nearly correct. The failure they share is named-entity contamination, which no metric detects and which flips up to 8 of 36 system pairs when ranking nine MT systems.
\footnote{\url{https://github.com/trotacodigos/rebeal}}
\end{abstract}

\section{Introduction}
\label{sec:intro}
 
For decades, machine translation (MT) has been evaluated by comparing system output to human-written reference translations. References are treated as gold, yet their quality varies with the translator and is difficult to verify~\citep{toral-sanchez-cartagena-2017-multifaceted}. Good references improve inter-annotator agreement in MT evaluation, especially when annotators are monolingual or lack domain expertise~\citep{ma-etal-2017-investigation}. In practice, however, they contain errors that bias judgment~\citep{fomicheva-specia-2016-reference, freitag-etal-2020-bleu}, and even error-free references provide evidence of merely one of many valid answers~\citep{qin-specia-2015-truly-exploring}. This was tolerable when MT was clearly inferior to human references.
 
That condition no longer holds. LLM-based MT can match or exceed human translation on standard benchmarks~\citep{kocmi-etal-2024-findings, kocmi-etal-2025-findings}, prompting calls for harder evaluation conditions~\citep{proietti-etal-2025-estimating, zouhar-etal-2026-generating}. Reference-free quality estimation (QE) has matured into a credible alternative and is increasingly used in place of, rather than alongside, reference-based metrics~\citep{agrawal-etal-2021-assessing, rei-etal-2022-cometkiwi, juraska-etal-2024-metricx, zerva-etal-2024-findings}. Both trends rest on an untested assumption that the information a reference provides is either redundant, when QE suffices, or reliable, when used. \citet{zouhar-bojar-2024-quality} show that vendor-level quality differences affect metric correlation, and the magnitude of the effect and the error types metrics are sensitive to remain uncharacterized.
 
Current data shows how far that assumption has already slipped. WMT annotates the human reference as one system among the MT submissions in its MQM campaigns, which makes reference quality measurable rather than assumed. Across the 2022 through 2024 campaigns \citep{kocmi-etal-2022-findings,kocmi-etal-2023-findings,kocmi-etal-2024-findings}, the reference ranks first in 5 of 6 language pairs in the earlier two years and in none of the three 2024 pairs, where the best system outscores it every time. Between 19\% and 87\% of reference segments carry at least one annotated error depending on the pair. The same annotations record which error types occur, which gives an empirical basis for deciding what a benchmark of reference corruption should contain. This grounds the contamination types in errors that occur, and it is not a premise the benchmark rests on. A metric that cannot separate a corrupted reference from a clean one is misreporting quality whatever the state of human translation; what has changed is how often that happens.
 
We build \mirrorbench (\textbf{Re}ference \textbf{B}ias \textbf{E}v\textbf{AL}uation) on that basis. For each segment we introduce one of seven contamination types into the reference, ranging from character-level noise to semantic distortions that preserve fluency, and measure how the score of each of eight widely used metrics changes across three language pairs. A robust metric should react to genuine corruption while ignoring superficial noise.

Our contributions are as follows.
\begin{itemize}
    \item We show that the human reference is no longer the best translation
available, ranking below the best system in every 2024 WMT MQM campaign, and we
characterize which errors it carries (\Cref{sec:practice}).

    \item We release \mirrorbench, a benchmark of seven controlled contamination
    types applied to human post-edited references in three language pairs
    (\Cref{sec:benchmark}).

    \item We benchmark eight metrics on it and find a blind spot every family
    shares, named entity contamination, which is hardest to detect at the
    intensity at which it occurs and which changes which system a metric ranks
    first (\Cref{sec:result}).
\end{itemize}
\section{Related Work}
\label{sec:related}

\paragraph{Reference Bias in MT Evaluation.}
Reference bias has been documented from multiple angles. \citet{fomicheva-specia-2016-reference} provide one of the earliest systematic examinations, showing through controlled experiments with multiple references that system rankings fluctuate significantly depending on reference choice. \citet{ma-etal-2017-investigation} extend this finding to monolingual annotation, reporting that annotators presented with low-quality references exhibit stronger bias and lower agreement. \citet{freitag-etal-2021-experts} note that even expert annotators are not immune to reference errors, which can lead to inconsistent judgments. On the automatic side, \citet{freitag-etal-2020-bleu} show that BLEU scores vary substantially when multiple references are used, and WMT shared tasks have reported similar instability for high-resource pairs with multiple references~\citep{freitag-etal-2021-results}. More recently, \citet{huang-etal-2024-lost} find that LLM-based evaluators over-rely on references and pay limited attention to the source segment. Together, these studies indicate that references are not neutral ground truth but an active source of bias.

\paragraph{Mitigating Reference Bias.}
Two complementary lines of work address this problem. The first improves the references themselves. \citet{thompson-post-2020-automatic} generate paraphrased references to reduce variance for surface-based metrics. \citet{noh-etal-2024-beyond} propose \textsc{ReSuMe}, which subtracts a reference baseline from metric scores so that translations exceeding reference quality can be evaluated more fairly. The second line removes references altogether. \citet{agrawal-etal-2021-assessing} demonstrate that a scaled multilingual MT model can match BLEU's correlation with human judgments using only the source. Dedicated QE metrics such as \textsc{Comet-Kiwi}~\citep{rei-etal-2022-cometkiwi, rei-etal-2023-scaling}, and \textsc{MetricX-24}~\citep{juraska-etal-2024-metricx} have since narrowed the gap further. \citet{Moosa2024MTRankerRM} introduce \textsc{MT-Ranker}, a ranking-based reference-free approach that remains stable under unreliable references. Both lines mitigate bias to some extent, but neither has been shown to consistently outperform reference-based evaluation.

\paragraph{Benchmarks for Reference Robustness.}
Few benchmarks directly target reference bias as the primary variable. Most prior efforts rely on naturally occurring variation, such as multiple human references~\citep{zouhar-bojar-2024-quality, zeng-etal-2024-towards} or limited paraphrastic perturbations~\citep{freitag-etal-2020-human, noh-etal-2024-beyond}, which conflate many error types and make it difficult to isolate the impact of specific corruption. Multilingual benchmarks such as those built around \textsc{xComet}~\citep{guerreiro-etal-2024-xcomet} evaluate metric robustness across languages and domains but do not manipulate reference quality. \mirrorbench addresses this gap with a structured perturbation framework spanning seven controlled corruption types across three language pairs, allowing fine-grained measurement of both the magnitude and the type-specific patterns of reference-induced bias.

\section{Reference Errors in Practice}
\label{sec:practice}

Reference-based evaluation assumes the reference is the better translation and that its content is correct. We test both assumptions against the WMT MQM campaigns from 2022 through 2024, covering 10 language pairs, of which 9 carry reference annotation. Detailed analysis is given in \Cref{appx:mqm} and \Cref{appx:entity}.

\begin{table}[t]
\centering\small
\setlength{\tabcolsep}{4pt}
\begin{tabular}{llccc}
\toprule
Year & Pair & Rank & \%Seg & Cov$_{\text{all}}$ / Cov$_{\text{maj}}$ \\
\midrule
2022 & En--De & 1, 3 / 16 & 30.3 & 19.7 / 40.3 \\
2022 & En--Ru & 1 / 16 & 40.5 & 29.8 / 17.1 \\
2022 & En--Zh$^{\dagger}$ & 1 / 13 & 86.2 & 46.6 / 29.7 \\
2023 & En--De & 1 / 13 & 58.9 & 28.0 / 31.6 \\
2023 & He--En & 1 / 13 & 30.2 & 37.6 / 32.2 \\
2023 & Zh--En & 12 / 16 & 86.7 & 30.6 / 39.1 \\
2024 & En--De & 6, 9 / 19 & 41.9 & 38.2 / 41.7 \\
2024 & En--Es & 8 / 15 & 18.6 & 61.8 / 50.0 \\
2024 & Ja--Zh & 2 / 15 & 31.8 & 47.7 / 58.0 \\
\midrule
\multicolumn{2}{l}{Median} & & 31.8 & 37.6 / 39.1 \\
\bottomrule
\end{tabular}
\caption{Human references in the WMT MQM campaigns. Rank places the reference among all annotated systems, with two entries where a campaign annotates two references. \%Seg is the share of reference segments carrying at least one error. Cov is the share of reference error annotations corresponding to a \mirrorbench contamination type, for all errors and for Major errors, counting category matches only. $\dagger$ marks the triple annotated release. Zh--En 2022 is omitted because its release carries no reference annotation.}
\label{tab:refrank}
\end{table}

\paragraph{The reference is no longer the best translation available.}
\Cref{tab:refrank} places each reference in the ranking of its own campaign. In 2022 and 2023 the reference ranks first in 5 of 6 pairs, which is the condition reference-based evaluation assumes. That condition fails in 2024. The reference ranks 2nd, 6th, 8th and 9th across the three 2024 campaigns, and the best system outscores it in every one. Reference quality also varies more than a single score conveys, since the share of reference segments carrying at least one annotated error ranges from 18.6--86.7\% across pairs.\footnote{Absolute MQM scores are not comparable across years because raters, guidelines and domains differ, so we report rank within campaign.}

\paragraph{The errors divide into forms that differ in what a metric would need to catch them.}
Named entity errors, which \Cat{Entity Swap} probes, account for 2.6\%, 7.4\% and 14.7\% of reference errors in the three releases that annotate them as a category, En$\rightarrow$Ru, En$\rightarrow$Es and He$\rightarrow$En, and are Major in 19\% of those cases at the lowest (\Cref{tab:entitygold}). \Cat{Content Addition} appears in every reference set and \Cat{Source Copy} in 7 of 9 (\Cref{tab:typerates}).\footnote{\Cat{Polarity Flip} and \Cat{Wrong Language} rest on span rules we do not validate, so we treat their observed rates as provisional.} Two criteria bound what the types cover. A type has to be isolatable as a single controlled corruption, which excludes mistranslation that resolves into no narrower form, and it has to be wrong against the source rather than a matter of degree, which excludes style, where no source check settles the judgment. The types therefore account for a median 37.6\% of reference error annotations, and Major coverage tracks that figure rather than exceeding it because style is almost never Major while the residual mistranslation often is.

\begin{table*}[t]
\centering\small
\setlength{\tabcolsep}{4pt}
\renewcommand{\arraystretch}{1.15}
\begin{tabular}{l p{5.2cm} p{3.0cm} p{3.6cm}}
\toprule
\textbf{Type} & \textbf{Description} & \textbf{Clean Reference} & \textbf{Perturbed Reference} \\
\midrule

\protect\circled{tPink!80}{A}\MinorNoiseCat & \textit{Ignore.} Characters in randomly selected tokens are reversed or deleted, producing typos and broken punctuation that leave meaning intact. & I \textit{have} a book. & I \textbf{ahve} a book. \\
\protect\circled{tPink!80}{A}\SourceCopyCat & \textit{Ignore.} Spans of the reference are replaced with the corresponding source fragments, simulating untranslated copy-through. & He bought \textit{a car}. & He bought \textbf{ein Auto}. \\
\midrule

\protect\circled{tGreen!70}{B}\WrongLangCat & \textit{Ignore.} The reference segment is replaced with a translation in a third language drawn from the other target languages in the benchmark. & He is tired. & \textbf{Está agotado.} \\
\protect\circled{tGreen!70}{B}\UnalignedCat & \textit{Ignore.} The reference is swapped with an adjacent segment from the same document, preserving style while the content no longer matches the source. & He is tired. & \textbf{He couldn't get up.} \\
\midrule

\protect\circled{tBlue!70}{C}\HalluCat & \textit{Check the source.} An adjacent segment is appended to the reference, mimicking hallucinated content that is topically relevant but unsupported by the source. & He is tired. & He is tired. \textbf{He couldn't get up.} \\
\protect\circled{tBlue!70}{C}\PolarityCat & \textit{Check the source.} Negation is added to or removed from the main predicate, inverting the polarity of the proposition. & The drug \textit{improved} symptoms. & The drug \textbf{did not improve} symptoms. \\
\protect\circled{tBlue!70}{C}\ArgSwapCat & \textit{Check the source.} Named entities are replaced with different entities of the same type, preserving the surrounding syntactic structure. & \textit{Vicente Siso}'s painting is exhibited. & \textbf{Pablo Picasso}'s painting is exhibited. \\
\bottomrule
\end{tabular}
\caption{The seven contamination types in \mirrorbench, grouped by category. Modified spans are shown in \textbf{bold}. The italic lead-in on each description gives what a robust metric should do with that type, following \Cref{subsec:design}: \protect\circled{tPink!80}{A} and \protect\circled{tGreen!70}{B} carry no translational meaning and should be disregarded, while \protect\circled{tBlue!70}{C} looks plausible and is caught only by cross-checking against the source. Real benchmark examples appear in \Cref{tab:real_ex}, and construction procedures in \Cref{appx:construction}.}
\label{tab:bias_type}
\end{table*}

\section{\mirrorbench}
\label{sec:benchmark}
\Cref{sec:practice} identifies the forms reference errors take and the two criteria that bound which of them a controlled probe can represent. We define seven contamination types on that basis.

\subsection{Design Principle}
\label{subsec:design}
We call a metric \textit{robust} when its score reflects the source--hypothesis relationship rather than agreement with the reference \citep{freitag-etal-2020-bleu,rei-etal-2020-comet}. A reference can supply contextual cues that are not recoverable from a single source sentence, and a reliable metric should exploit them. When the reference is itself corrupted, the metric should discount the corrupted portion and prefer a clean reference over a noisy one.
 
The corruptions we probe take two forms. Some content should be \textit{ignored} because it carries no translational meaning, such as typographical noise or fragments left in the wrong language. Other content should be \textit{verified against the source} because it looks plausible but contradicts what the source actually says. A robust metric must handle both, and the contamination types defined below test each behavior in isolation.

\subsection{Contamination Types}
\label{subsec:perturbations}

\Cref{tab:bias_type} defines seven contamination types grouped into three
categories. \circled{tPink!80}{A} \textbf{Surface noise} alters characters
without changing meaning. \circled{tGreen!70}{B} \textbf{Irrelevant content}
substitutes or inserts material unrelated to the source segment.
\circled{tBlue!70}{C} \textbf{Hallucination} produces references unfaithful to
the source, either by adding unsupported content (extrinsic) or by contradicting
it (intrinsic)~\citep{raunak-etal-2021-curious, guerreiro-etal-2023-hallucinations}.
The types differ in how far observed reference errors support them, and
\Cref{appx:mqm} gives the observed rate of each together with the evidence
behind it.

\subsection{Dataset Construction}
References start from human post-edited translations, which \Cref{sec:practice} shows to be of high quality without being error free. Each item pairs a clean and a contaminated version of the same segment, so any residual error in the source reference enters both conditions and does not confound the difference we measure. Most contamination follows rule-based procedures that give precise control over intensity and location. The two categories requiring finer semantic control, \Cat{Polarity Flip} and \Cat{Entity Swap}, are generated by prompting \mGpt (\verb|gpt-5_5-2026-04-23|), with full procedures and prompts in \Cref{appx:construction,appx:prompts}. Some types such as \Cat{Minor Noise} and \Cat{Source Copy} are graded into three contamination levels at 30, 60, and 90 percent of the reference, while \Cat{Entity Swap} levels follow within-language tertiles of entity count.
 
We build \mirrorbench on WMT24++~\citep{deutsch-etal-2025-wmt24}, which provides high-quality human post-edited translations. The source side contains 961 segments across four domains (\textit{social}, \textit{news}, \textit{speech}, \textit{literary}). We reduce this to 705 segments per language pair to balance domain proportions and exclude segments too short to contaminate meaningfully, yielding 4{,}935 contaminated samples per pair across the seven contamination types. Detailed statistics appear in \Cref{appx:stats}.

\subsection{Language Pair}
We cover English$\rightarrow$ \{German, Chinese, Korean\} (en$\rightarrow$de, en$\rightarrow$zh, en$\rightarrow$ko), chosen to vary in script and typology. German shares the Latin script and is morphologically close to English. Chinese is logographic and written without whitespace, so a character-level edit alters more meaning-bearing units per edit than it does in an alphabetic script; the effect is script-specific rather than a matter of segmentation, since Korean uses spaces and \Cat{Minor Noise} raises TTR by $0.376$ in Chinese against $0.165$ in Korean (\Cref{tab:stats-combined}). Korean is agglutinative and follows SOV order, introducing morphological richness and reordering. This contrast lets us examine whether metric robustness depends on language rather than on the contamination type alone.

\subsection{Human Quality Check}
We validate \mirrorbench through a pairwise preference study. For each language pair we sample 300 instances stratified by contamination type and contamination level. Annotators see the source segment together with two candidate translations (the clean and the contaminated reference) in randomized order, and choose which one they prefer or mark the pair as tied. Each language pair is judged by two native speakers and is double-annotated. Across all pairs, clean references are preferred over contaminated ones in 99.8\% of cases on average. The remaining 0.2\% are ties rather than preferences for the contaminated version.

\begin{table*}[t]
\centering
\small
\setlength{\tabcolsep}{4pt}
\resizebox{\linewidth}{!}{
\renewcommand{\arraystretch}{1.15}
\begin{tabular}{l c c c c c c c | c}
\toprule
 & \Cat{Minor Noise} & \Cat{Source Copy} & \Cat{Wrong Lang} & \Cat{Unaligned Sent} & \Cat{Content Add} & \Cat{Polarity Flip} & \Cat{Entity Swap} & \textbf{Avg} \\
 & $\nu$  /  $\Delta$ & $\nu$  /  $\Delta$ & $\nu$  /  $\Delta$ & $\nu$  /  $\Delta$ & $\nu$  /  $\Delta$ & $\nu$  /  $\Delta$ & $\nu$  /  $\Delta$ & $\nu$  \\
\midrule
\multicolumn{9}{l}{\textit{Reference-based}} \\
\chrf & .00 / +19.7 & .01 / +25.5 & .00 / +40.0 & .00 / +26.9 & .00 / +13.0 & .06 / +5.33 & \sMax{.98 / -43.3} & .15\\
\bertscore & .00 / +.094 & .00 / +.100 & .00 / +.060 & .04 / +.066 & .00 / +.023 & .05 / +.011 & \sMax{.95 / -.037} & .15 \\
\unite & .01 / +.741 & .00 / +.959 & .04 / +.490 & .06 / +.934 & .03 / +.567 & .02 / +.395 & \sHi{.65 / -.077} & .12 \\
\xcomet & .02 / +.326 & .01 / +.368 & \sLow{.12 / +.176} & .04 / +.463 & .03 / +.123 & .07 / +.073 & \sMid{.27 / +.060} & .08 \\
\metricx & .01 / +8.10 & .01 / +9.56 & \sLow{.14 / +.783} & .04 / +7.11 & .00 / +7.23 & .04 / +1.16 & \sMid{.23 / +1.18} & .07 \\
%\midrule
%%\multicolumn{9}{l}{\textit{Reference-free (QE, $\Delta$ floors at zero by construction)}} \\
%\uniteqe & .25 / -- & .25 / -- & .25 / -- & .24 / -- & .25 / -- & .24 / -- & .25 / -- & .25  \\
%\xcometkiwi & .20 / -- & .19 / -- & .19 / -- & .18 / -- & .20 / -- & .19 / -- & .19 / -- & .19\\
%\metricxqe & .17 / -- & .18 / -- & .17 / -- & .21 / -- & .17 / -- & .18 / -- & .19 / -- & .18 \\
\bottomrule
\end{tabular}
}
\caption{Violation Rate ($\nu$, lower is better) and Robustness Gap ($\Delta$,
higher is better) averaged over all language pairs, with per-language tables in
\Cref{tab:lang-en-de} to \Cref{tab:lang-en-zh}. Cells are shaded by $\nu$ magnitude, and negative
$\Delta$ marks metric subversion, where the contaminated reference scores above
the clean one on average. Reference-free QE metrics are omitted. They never
consult the reference, so a clean and a contaminated reference are identical
input to them and neither measure is defined. \Cref{subsec:qe} uses that
invariance as a fixed baseline against which the movement of a reference-based
score is read.}
\label{tab:vr_rg_all}
\end{table*}

\section{Experiment Setup}
\label{sec:setup}

\subsection{MT Metrics}
We evaluate eight metrics spanning three families. Three criteria govern the
selection. We take at least one metric per scoring paradigm, so that a failure can
be attributed to how a metric is built rather than to a particular model: n-gram
overlap, contextual embedding similarity, and learned regression over MQM
judgments. Within the learned paradigm we take the two strongest reference-based
metrics of the WMT24 metrics shared task, and we retain \chrf and \bertscore
because both remain in wide use for system comparison despite being superseded on
that task. Finally, every learned metric is paired with a QE counterpart trained
on the same backbone, which is what makes the fixed-baseline comparison in
\Cref{subsec:qe} possible.
\textbf{Surface-based metrics} compute n-gram overlap on character or word units. We use \chrf~\citep{popovic-2015-chrf} (for Chinese, and \chrfpp for German and Korean) implemented via SacreBLEU~\citep{post-2018-call}. \textbf{Learned reference-based metrics} score candidate translations against the reference using neural models, and we include \bertscore~\citep{zhang-2020-bertscore}, \unite~\citep{wan-etal-2022-unite}, \xcomet~\citep{guerreiro-etal-2024-xcomet}, and \metricx~\citep{juraska-etal-2024-metricx}. \textbf{Reference-free (QE) metrics} score translations using only the source. We include \uniteqe, \xcometkiwi~\citep{rei-etal-2023-scaling}, and \metricxqe, the QE variants of the corresponding reference-based models. QE metrics assign the same score whether the reference is clean or contaminated, since they do not consult it. They are a fixed baseline for reading how far a reference-based score has moved, not an ideal to match: a metric that uses the reference should outperform one that ignores it. Model and tokenizer details for each metric are listed in \Cref{appx:metric-config}.

\subsection{Robustness Measures}
We evaluate whether a metric consistently distinguishes a clean reference from a contaminated counterpart. Let $h_i$ denote the hypothesis translation, $r_i$ the clean post-edited reference, and $\tilde{r}_i$ a contaminated reference. We fix $h_i$ across conditions and vary the reference between $r_i$ and $\tilde{r}_i$. Let $M(h, r)$ denote the metric score, with the source omitted from notation. For lower-is-better metrics (\textsc{MetricX} series, \ter), we negate scores so that higher values always indicate better quality.

\paragraph{Violation Rate ($\nu$).}
The proportion of segments where the contaminated reference receives a higher score than the clean reference. A robust metric should have $\nu \approx 0$.
\begin{equation}
    \nu = \frac{1}{N} \sum_{i=1}^{N} \mathbbm{1}
    \big[ M(h_i, r_i) < M(h_i, \tilde{r}_i) \big]
\end{equation}

\paragraph{Robustness Gap ($\Delta$).}
The average difference between clean and contaminated scores. Positive values indicate that the metric assigns higher quality to the clean reference, so a robust metric should have $\Delta > 0$.
\begin{equation}
    \Delta = \frac{1}{N} \sum_{i=1}^{N}
    \big[ M(h_i, r_i) - M(h_i, \tilde{r}_i) \big]
\end{equation}

The gap is reported in each metric's native scale, and cross-metric comparison uses the violation rate ($\nu$). We report $\Delta$ with $95\%$ confidence intervals estimated from $2{,}000$ bootstrap samples.

The violation rate captures the \textit{frequency} of robustness violations, while the gap ($\Delta$) captures their \textit{magnitude}. Together they characterize how reliably a metric distinguishes clean from contaminated references when scoring the same hypothesis.
\section{Results}
\label{sec:result}
We report robustness at the segment level in \Cref{subsec:robustness} through \Cref{subsec:dose}, and at the system level in \Cref{subsec:ranking}.

\subsection{Robustness Measures}
\label{subsec:robustness}

\paragraph{Learned metrics are more robust.}
Averaged across the seven contamination categories, \xcomet ($\bar{\nu}=0.08$) and \metricx ($\bar{\nu}=0.07$) achieve the lowest violation rates, while \chrf and \bertscore sit higher ($\bar{\nu}=0.15$) (\Cref{tab:vr_rg_all}). The aggregate hides where each family fails. \chrf and \bertscore stay near zero on every category but collapse on \Cat{Entity Swap} ($\nu=0.98$ and $0.95$), which alone drives their average. \xcomet and \metricx keep $\nu$ below $0.15$ everywhere except \Cat{Entity Swap} ($0.27$ and $0.23$, the lowest of any reference-based metric) and \Cat{Wrong Language} ($0.12$ and $0.14$, a category surface metrics catch trivially).

\paragraph{Entity Swap raises scores above the clean reference.}
\Cat{Entity Swap} is the only contamination that drives surface metrics to assign scores \emph{higher} than the clean reference. We call this \emph{metric subversion}: the contaminated reference outscores the clean one, so the metric does not merely miss the corruption but rewards it, and a negative $\Delta$ captures it (\Cref{tab:vr_rg_all}). \chrf, for instance, reaches $\nu=0.98$ and $\Delta=-43.3$, meaning the contaminated reference is preferred over the clean one in nearly every segment, with a mean uplift of 43 points. Despite the better performance of \xcomet and \metricx in this regard ($\nu=0.27$ and $0.23$ respectively), one in four segments still receives a higher score when contaminated.

\paragraph{Learned metrics tolerate an unrelated language.}
Surface metrics detect \Cat{Wrong Language} almost universally (\chrf $\Delta=+40.0$, $\nu=0.00$), while \xcomet and \metricx show elevated violation rates ($0.12$ and $0.14$), an order of magnitude higher than for most other contamination. Their gaps are also small relative to those on \Cat{Unaligned Sent} or \Cat{Source Copy} (\metricx $\Delta=0.78$ vs.\ $7.11$), indicating that learned metrics tolerate references in unrelated languages more readily than same-language semantic divergence, likely a side effect of their multilingual backbones (XLM-R, mT5). The effect also varies by language pair (\Cref{tab:lang-en-de,tab:lang-en-ko,tab:lang-en-zh}).

\begin{table}[t]
\centering
\small
\setlength{\tabcolsep}{4pt}
\renewcommand{\arraystretch}{1.15}
\resizebox{\linewidth}{!}{
\begin{tabular}{l l ccc}
\toprule

 &  & $L{=}0.3$ & $L{=}0.6$ & $L{=}0.9$ \\
 &  & $\nu$ / $\Delta$ & $\nu$ / $\Delta$ & $\nu$ / $\Delta$ \\
\midrule
\multirow{5}{*}{\Cat{Minor Noise}} & \chrf & .00 / +13.0 & .00 / +20.3 & .00 / +25.7 \\
 & \bertscore & .01 / +.068 & .00 / +.094 & .00 / +.119 \\
 & \unite & .02 / +.555 & .01 / +.784 & .00 / +.886 \\
 & \xcomet & .03 / +.214 & .01 / +.295 & .00 / +.468 \\
 & \metricx & .01 / +6.19 & .01 / +7.92 & .00 / +10.2 \\
\midrule
\multirow{5}{*}{\Cat{Source Copy}} & \chrf & .02 / +17.6 & .01 / +26.0 & .01 / +33.1 \\
 & \bertscore & .00 / +.072 & .00 / +.101 & .00 / +.126 \\
 & \unite & .01 / +.797 & .00 / +1.05 & .00 / +1.03 \\
 & \xcomet & .02 / +.240 & .01 / +.399 & .00 / +.466 \\
 & \metricx & .01 / +8.07 & .01 / +10.0 & .01 / +10.6 \\
\midrule
\multirow{5}{*}{\Cat{Entity Swap}} & \chrf & \sMax{\textbf{1.00 / -49.3}} & \sMax{.97 / -43.3} & \sMax{.96 / -32.5} \\
 & \bertscore & \sMax{\textbf{.98 / -.043}} & \sMax{.95 / -.038} & \sMax{.92 / -.027} \\
 & \unite & \sMax{\textbf{.77 / -.178}} & \sHi{.66 / -.051} & \sMid{.39 / +.065} \\
 & \xcomet & \sMid{\textbf{.38 / +.015}} & \sMid{.25 / +.070} & .09 / +.129 \\
 & \metricx & \sMid{\textbf{.30 / +.703}} & \sMid{.23 / +1.22} & \sLow{.12 / +2.07} \\
\bottomrule
\end{tabular}
}
\caption{Per-level robustness for the three multi-level categories ($L \in \{0.3, 0.6, 0.9\}$), averaged across all language pairs. Cells are shaded by $\nu$ magnitude.}
\label{tab:dose-response}
\end{table}

\subsection{The Smallest Swap Is the Most Dangerous}
\label{subsec:dose}

We further examine \Cat{Entity Swap} by varying contamination intensity
(\Cref{tab:dose-response}). Unlike other categories, the corruption becomes
\emph{less} visible to learned metrics as fewer entities are hallucinated.
\xcomet $\nu$ rises from $0.09$ at $L=0.9$ to $0.38$ at $L=0.3$, and \metricx
from $0.12$ to $0.30$. A single mismatched entity slips past \xcomet in $38\%$ of
segments and \metricx in $30\%$, despite being obvious to a human reader, while
surface metrics are subverted at every level.

The low-intensity regime is the one that matches observation. Among the reference
segments that carry an annotated named entity error, $87\%$ carry exactly one
(\Cref{sec:practice}), so an affected reference corresponds to the lowest
contamination level rather than the highest. Those errors reach Major severity in
$19\%$ of cases at the lowest across the three releases that annotate them
separately. Metrics are weakest in exactly the regime where observed reference
errors sit.

\begin{figure*}
    \centering
    \includegraphics[width=0.90\linewidth]{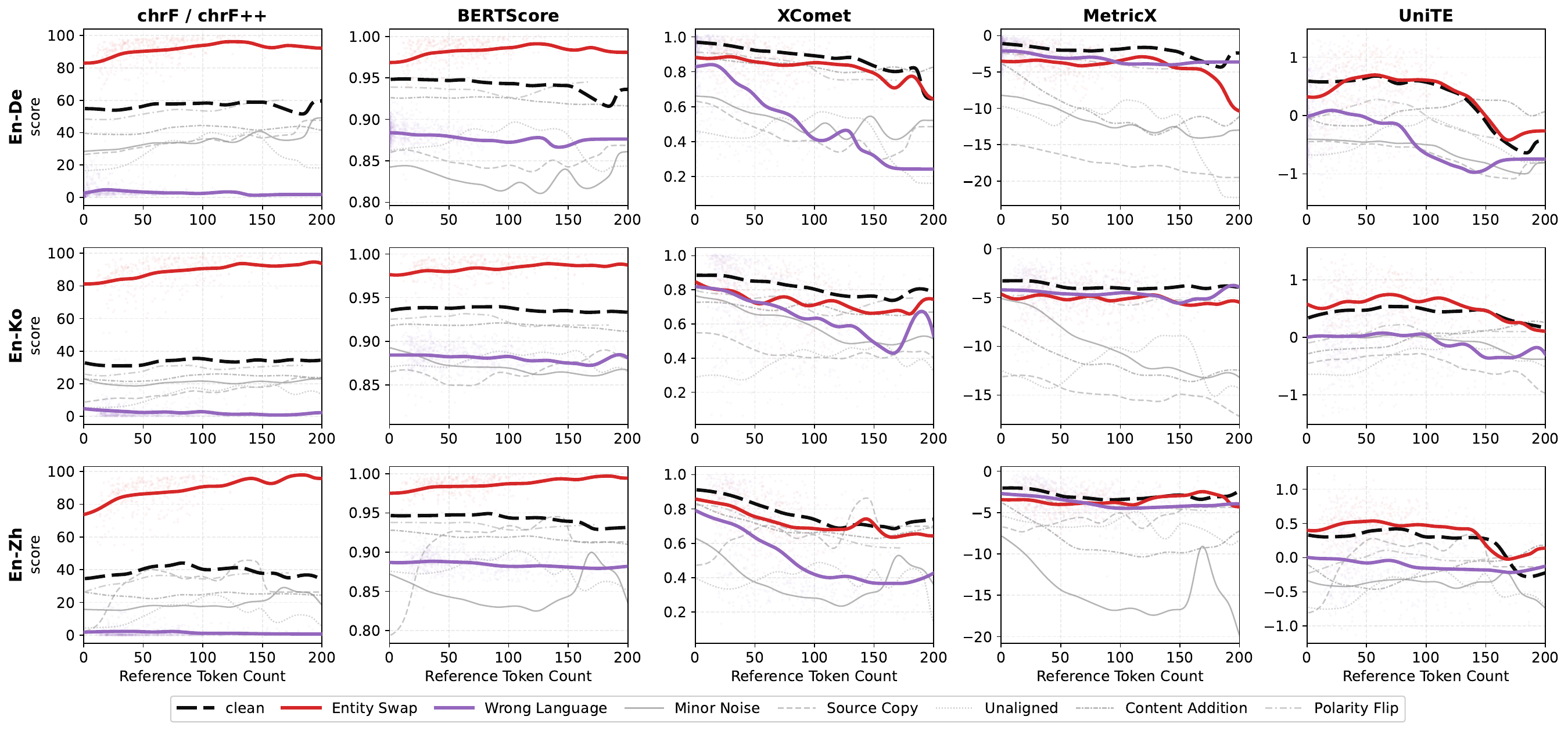}
    \caption{Metric scores on contaminated references as a function of reference token count. The black dashed line marks the score on the clean reference. \protect\Cat{Entity Swap} (\textcolor{red!80!black}{red}) and \protect\Cat{Wrong Language} (\textcolor{violet!80!black}{purple}) are foregrounded, while other contamination types appear in muted gray.}   
    \label{fig:length}
\end{figure*}

\begin{table}[t]
\centering\small
\setlength{\tabcolsep}{4pt}
\renewcommand{\arraystretch}{1.1}
\begin{tabular}{l ccc}
\toprule
\textbf{Metric} & \textbf{en--de} & \textbf{en--zh} & \textbf{en--ko} \\
\midrule
\multicolumn{4}{l}{\EntitySwapCat} \\
\chrf       & 0.61 / 7 & 0.56 / 8 & 0.67 / 6 \\
\bertscore  & 0.67 / 6 & 0.61 / 7 & 0.72 / 5 \\
\unite      & 0.78 / 4 & 0.72 / 5 & 0.78 / 4 \\
\xcomet     & 0.89 / 2 & 0.83 / 3 & 0.89 / 2 \\
\metricx    & 0.89 / 2 & 0.83 / 3 & 0.94 / 1 \\
\midrule
\multicolumn{4}{l}{\WrongLangCat} \\
\chrf       & 1.00 / 0 & 1.00 / 0 & 0.94 / 1 \\
\bertscore  & 0.94 / 1 & 1.00 / 0 & 0.94 / 1 \\
\unite      & 0.83 / 3 & 0.78 / 4 & 0.83 / 3 \\
\xcomet     & 0.83 / 3 & 0.78 / 4 & 0.83 / 3 \\
\metricx    & 0.83 / 3 & 0.83 / 3 & 0.78 / 4 \\
\bottomrule
\end{tabular}
\caption{System-level effect of reference contamination. Each cell gives Kendall's $\tau$ against the clean-reference ranking of 9 MT systems on WMT24++, followed by the number of the 36 system pairs whose order flips. Lower $\tau$ and more flips indicate a ranking the contamination has disturbed.}
\label{tab:ranking}
\end{table}

\subsection{Contamination Changes Which System Ranks First}
\label{subsec:ranking}

The measures so far are segment-level, and metrics are used to compare systems.
We rank 9 MT systems on WMT24++ (en$\rightarrow$de, en$\rightarrow$zh,
en$\rightarrow$ko) under clean and contaminated references and report Kendall's
$\tau$ against the clean ranking together with the number of the 36 system pairs
whose order flips (\Cref{tab:ranking}).

Rankings hold for five of the seven contamination types, where $\tau \geq 0.92$
and at most 2 of 36 pairs flip. The two exceptions follow the segment-level
pattern. Under \Cat{Entity Swap}, \chrf and \bertscore break down with $\tau$
falling to $0.56$ and 8 of 36 pairs flipped, and \xcomet and \metricx are affected
as well at $\tau \approx 0.85$ with 2 to 3 flips. Under \Cat{Wrong Language} the
opposite holds, where \chrf is barely moved at $\tau \approx 1.0$ while the learned
metrics shift. The failures we report are not confined to absolute scores. They
change which system a metric ranks first.

\subsection{Sensitivity to Reference Length}
\label{subsec:length}

\Cref{fig:length} plots metric scores on contaminated references against reference token count, smoothed with a Gaussian kernel over per-segment scores. Surface metrics remain flat across length on most categories, while \xcomet and \unite do not. We focus on \Cat{Entity Swap} and \Cat{Wrong Language}, the two categories where this contrast is most informative. On \Cat{Entity Swap}, \chrf and \bertscore score contaminated references \emph{above} clean across the full length range in all three language pairs, and \xcomet and \metricx stay close to or slightly below clean without recovering at longer lengths. On \Cat{Wrong Language}, \xcomet instead shows a clear length gradient. Contaminated scores sit close to the clean baseline on short references and fall further below it as length grows. The effect is sharpest on en--de, where the \xcomet score on a contaminated reference starts at $0.83$ on the shortest segments and falls to $0.24$ on the longest, with short-reference scores barely distinguishable from clean. \metricx, however, remains weak on this category throughout.

\begin{figure*}[t]
    \centering
    \includegraphics[width=0.88\linewidth]{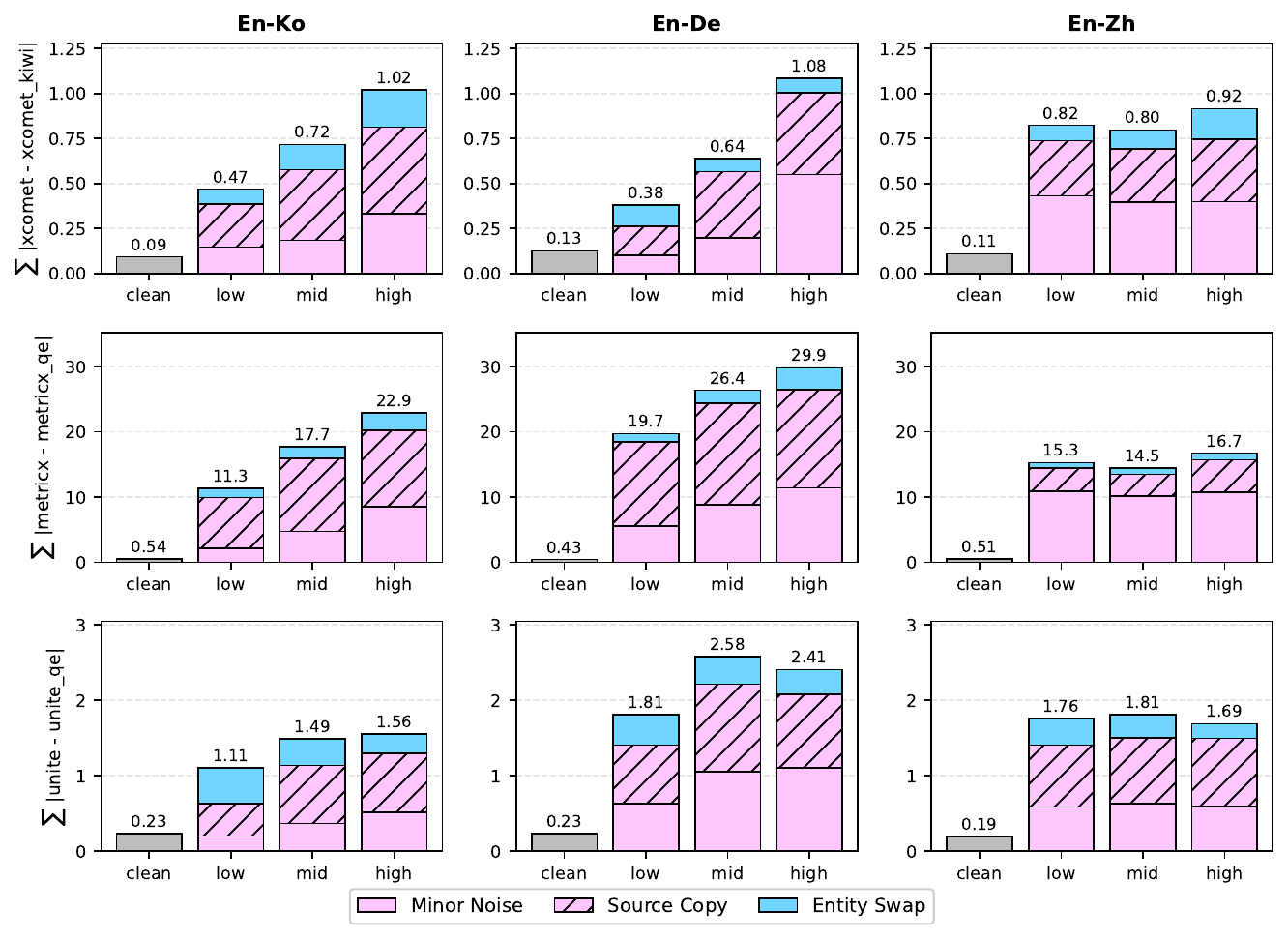}
    \caption{Movement of each reference-based metric away from its QE counterpart, by language pair and contamination level. QE is invariant to reference contamination, so the bars measure the reference-based score. Bars are stacked across the three multi-level categories and colored by their typology. The grey bar at \emph{clean} marks the baseline disagreement on identical input.}
    \label{fig:qe_gap}
\end{figure*}

\subsection{Reference Sensitivity Against a Fixed Baseline}
\label{subsec:qe}

QE scores depend only on the source and the hypothesis, so they are invariant to
reference contamination by construction. That invariance makes them a fixed
baseline against which the movement of a reference-based score can be read. We
measure the segment-level absolute gap $|M_\mathrm{ref} - M_\mathrm{QE}|$ between
each reference-based metric $M$ and its QE counterpart, average it across the
three multi-level categories, and report it by contamination level. Any change in
the gap is a change in $M_\mathrm{ref}$.

\paragraph{Entity Swap moves the reference-based score least.}
The \Cat{Entity Swap} segments barely rise above the clean baseline regardless of
how many entities are swapped. On \xcomet the gap moves only $-0.05$ to $+0.12$
from clean to high level across the three language pairs. The reference-based
score therefore barely moves whether one entity is swapped or many, which matches
the flat dose response in \Cref{subsec:dose}.

\paragraph{Surface noise moves it most.}
The \Cat{Source Copy} and \Cat{Minor Noise} segments grow substantially with
contamination level. On \xcomet for en--ko, the \Cat{Source Copy} contribution
rises from $0.24$ at low to $0.48$ at high. The pattern is monotonic on Korean and
German but saturates on Chinese, where even low-level contamination already
produces a gap of $0.31$, comparable to the high-level gaps of the other two
languages.

\begin{table}[t]
\centering\small
\setlength{\tabcolsep}{4pt}
\begin{tabular}{lcccc}
\toprule
\textbf{Metric} & \textbf{clean} & \textbf{flawed} & $d$ & $\rho$ with MQM \\
\midrule
\chrf        & 59.9 & 52.2 & 0.47 & \phantom{$-$}.14 $\rightarrow$ $-$.16 \\
\xcomet      & .961 & .910 & 0.61 & \phantom{$-$}.32 $\rightarrow$ \phantom{$-$}.25 \\
\xcometkiwi  & .963 & .895 & 0.69 & \phantom{$-$}.34 $\rightarrow$ \phantom{$-$}.27 \\
\bottomrule
\end{tabular}
\caption{Scoring refB against refA on WMT24 en$\rightarrow$de, over the 288
segments where refB carries no annotated error. The \textbf{clean} column covers
the 73 segments whose refA is also error-free and \textbf{flawed} the remaining
215. $d$ is the standardised difference between the two groups and $\rho$ the
segment-level Spearman correlation with MQM. \xcometkiwi is the QE counterpart of
\xcomet; it never sees refA and separates reference corruption from segment
difficulty.}
\label{tab:natural}
\end{table}

\subsection{Naturally Flawed References Reproduce the Pattern}
\label{subsec:natural}

Every contamination so far is one we introduced. WMT24 en$\rightarrow$de
annotates two human references, refA and refB, which lets us repeat the
comparison on errors nobody planted. We treat refB as a translation and score it
against refA.

We keep the 288 segments where refB carries no annotated error, so the
translation being scored is sound in all of them. Of those, 73 have a clean refA
and 215 a flawed one (\Cref{tab:natural}). In the second group the translation is
right and the reference is wrong, which is the condition a robust metric should
survive.

\chrf reverses sign. Once refA is flawed it correlates with human judgment at
$-.16$, which is to say it begins to prefer the segments where the reference is
wrong. \xcomet also drops, but \xcometkiwi never sees refA and drops by as much,
so that loss is a property of the segments rather than of the reference. \metricx
keeps its correlation on the same test ($\rho = .26 \rightarrow .24$). The
ordering matches \Cref{tab:vr_rg_all}: the metric a synthetic entity swap
subverts most is the one naturally flawed references disturb most. This covers
one language pair and one translation, so we read it as corroboration of the
controlled result rather than as an independent measurement of it.

\section{Can the Blind Spot Be Closed?}
\label{sec:mitigation}

The benchmark diagnoses rather than repairs, and a natural question is whether
the \Cat{Entity Swap} failure can be removed by combining scores a practitioner
already has. We test three combinations of a learned metric with its
same-backbone QE counterpart.

Averaging the two leaves $\nu$ unchanged by construction, since the QE term is
identical under the clean and the contaminated reference and cancels from the
comparison. Taking the minimum does lower $\nu$ on \Cat{Entity Swap}, from $0.38$
to $0.21$ for \xcomet, but it reaches that figure by selecting the QE score in
55--80\% of segments, which is to say by discarding the reference in most of
them. Treating the gap $|M_\mathrm{ref} - M_\mathrm{QE}|$ as a contamination
detector works where the contamination is visible on the surface, with AUROC
between $0.85$ and $0.88$ for \Cat{Source Copy} and \Cat{Unaligned Sent}, and is
at chance for \Cat{Entity Swap} on \xcomet (AUROC $0.52$).

Post-hoc combination therefore does not reach this failure. The signal a combined
score would need is absent from both terms, which places the fix in training
rather than in aggregation. \Cref{sec:model_behavior} indicates that the signal
is learnable: a translation model on the same backbone discounts a swapped entity
almost entirely ($\Delta_{\textsc{ter}} = 1.4$ against 23--25 for the other
contamination types) while the metrics built on that backbone do not. Two
recommendations follow. Entity consistency between source and hypothesis is a
candidate training objective for reference-based metrics, and reporting a
reference-based score beside its QE counterpart costs nothing and exposes the
cases where the two disagree.

\section{Conclusion}

Reference-based evaluation assumes the reference is correct and better than the
systems it scores. Neither assumption holds in the WMT MQM campaigns we examine,
where the human reference ranks below the best system in every 2024 language pair
and carries annotated errors in up to 87\% of its segments. \mirrorbench measures
what metrics do once that assumption fails.

Metrics agree on easy contamination and diverge on hard contamination, and each
family diverges in its own way. \chrf and \bertscore show the same failure
regardless of segment length, so substantial contamination is masked by the
surrounding text once a segment is long enough. \xcomet and \metricx fail
elsewhere. Multilingual capacity, the property that makes them strong, leaves them
unable to treat a reference written in another language as wrong.

One failure is shared. No metric detects a swapped named entity, the corruption is
hardest to catch at the single-entity intensity at which it occurs in real
references, and it changes which system a metric ranks first. A translation model
given the same contaminated reference barely moves its output, an order of
magnitude less than for the other six types. That places the failure in what
metrics attend to rather than in what their backbones can represent. Metrics and
translation models are increasingly built on the same multilingual encoders and
trained for different objectives. A metric that has both the source and the
reference in front of it can learn when to weigh one against the other rather than
treat them as a single composite target.

\section*{Limitations}

\paragraph{Coverage of contamination types.}
The seven types account for a median 37.6\% of reference error annotations in
\Cref{sec:practice}. What falls outside is not arbitrary. Style and fluency errors
supply the largest share and almost never reach Major severity, while
mistranslation that resolves into no narrower form is less frequent and often
Major, and it accounts for most of the Major reference errors we do not probe.
Neither is reachable by our design, since a probe that isolates one corruption at
a time cannot represent a judgment of degree or an error defined by the absence of
narrower structure. Document-level failures are outside the segment-level scope of
\mirrorbench for the same reason.

\paragraph{One contamination at a time.}
Each \mirrorbench item applies a single contamination type, which is what lets a
specific failure be attributed to a specific metric. Real references are not so
orderly. A segment can carry a swapped entity inside an awkward sentence, or an
untranslated fragment alongside an omission, and the MQM annotations in
\Cref{sec:practice} contain segments with several errors at once. Whether types
mask or amplify one another, and whether a metric that catches each in isolation
still catches them in combination, is outside what this benchmark can answer.

\paragraph{English-source language pairs.}
\mirrorbench covers three language pairs (en$\rightarrow$de, en$\rightarrow$zh,
en$\rightarrow$ko), all with English as the source. The patterns we report may
interact with the dominance of English in metric training data, and whether they
generalize to non-English source languages or to typologically more distant pairs
(e.g., between two low-resource languages) is an open question. The observational
evidence in \Cref{sec:practice} does not close this gap, since the WMT MQM
campaigns cover 7 language pairs, none of which is en$\rightarrow$ko. They
establish that these corruption types occur in human references across languages,
not that they occur at any particular rate in Korean. Observed rates vary by a
factor of 5 across pairs for a single type, so we read them as evidence of
existence rather than of frequency. The restriction follows from what a clean
baseline requires. \mirrorbench starts from post-edited references, and WMT24++
is the only source of them at the scale we need; it covers en$\rightarrow$X only.
The non-English-source references in \Cref{sec:practice} carry annotated errors
in 31--87\% of their segments and cannot serve as a clean starting point. None of
the contamination procedures depends on the source language, so the benchmark
extends as soon as post-edited references for non-English sources exist.

\paragraph{LLM-generated contamination for two categories.}
\Cat{Polarity Flip} and \Cat{Entity Swap} are generated by prompting an LLM. While this approach gives us fine-grained semantic control that rule-based procedures cannot, the resulting perturbations may carry stylistic regularities specific to the generating model, and may differ from how analogous errors appear in human-written or system-generated references.

\paragraph{Reliance on spaCy for entity detection.}
Our \Cat{Entity Swap} construction depends on spaCy for NER, whose coverage is known to vary across languages and entity types, but the model may miss entities that a more capable tagger or a human annotator would identify. As a result, the perturbed set may contain fewer entity replacements than the underlying text actually permits. To avoid penalizing this asymmetry, segments in which spaCy detects no entities are excluded from the corresponding analyses rather than treated as zero-entity cases.

\paragraph{Potential data leakage in the MT model.}
The generation-side experiments use \mQwen on segments drawn from WMT24++, which may overlap with the model's pre-training data. The exact training cutoff is not publicly documented, and we cannot rule out that the model has seen some of the source sentences or even references during pre-training. We note, however, that our analysis does not depend on absolute translation quality. The measurements isolate how much a model's output shifts when the reference shown in the prompt is contaminated. Any prior exposure to the source would affect both conditions equally and is therefore absorbed into the baseline.

\paragraph{Scope of the generation analysis.}
The generation analysis in \Cref{sec:model_behavior} covers one translation model and one prompting setup, and it supports the asymmetry we report rather than establishing it across systems. Supplying a reference at inference time is also not a deployment scenario. Post-editing poses the same question in a deployed form and would test whether a model overrides a corrupted entity when revising an existing translation rather than when conditioned on a gold reference.
\section*{Acknowledgment}
We acknowledge the use of AI assistants (\textsc{Claude Opus 4.7}) for writing refinement and code review during paper preparation.

This work was supported by the G-LAMP Program of the National Research Foundation of Korea (NRF) grant funded by the Ministry of Education (No. RS-2025-25441317); the Ministry of Science and ICT (MSIT) and the National IT Industry Promotion Agency (NIPA) through the Advanced GPU Utilization Support Program (02-26-01-0499); the NRF grant funded by the MSIT (RS-2025-16071992). This work was supported by Korea Internet \& Security Agency(KISA) grant funded by the Korea government(PIPC) (No.RS-2026-25526342, Development of Technologies for Preventing Sensitive Information Inference and Risk Assessment in Foundation Model Operations). This research was supported by Culture, Sports and Tourism R\&D Program through the Korea Creative Content Agency grant funded by the Ministry of Culture, Sports and Tourism in 2026 (Project Name: Develop AI agent technology to connect knowledge through public cultural facility-based discussion and communication, Project Number: RS-2026-25520645).

\section*{Ethics Statement}
Our study involves human annotators. Participation was voluntary, and annotators were compensated at a fair market rate. All annotations were anonymized, and no personally identifiable information was collected.

Licenses of artifacts. All datasets, models, and software used in this study are publicly available for research purposes. The WMT24++ test set is distributed under CC-BY 4.0; the automatic metrics are released under Apache 2.0 and MIT. \mGpt is used via the OpenAI API under OpenAI's terms of service. Our benchmark and code will be released under CC-BY 4.0 and MIT, respectively.

\textbf{Intended use.} Our use of the WMT24++ test set and automatic metrics is consistent with their intended use for MT evaluation research. The annotations and datasets we release are intended for research use only, consistent with the access conditions of the source data.

\bibliography{anthology-2, arxiv/arxiv}

@inproceedings{snover-etal-2006-study,title = "A Study of Translation Edit Rate with Targeted Human Annotation",author = "Snover, Matthew and Dorr, Bonnie and Schwartz, Rich and Micciulla, Linnea and Makhoul, John",booktitle = "Proceedings of the 7th Conference of the Association for Machine Translation in the Americas: Technical Papers",month = aug # " 8-12",year = "2006",address = "Cambridge, Massachusetts, USA",publisher = "Association for Machine Translation in the Americas",url = anth # {2006.amta-papers.25/},pages = "223--231"}

@inproceedings{zouhar-bojar-2024-quality,
    title = "Quality and Quantity of Machine Translation References for Automatic Metrics",
    author = "Zouhar, Vil{\'e}m  and
      Bojar, Ond{\v{r}}ej",
    editor = "Balloccu, Simone  and
      Belz, Anya  and
      Huidrom, Rudali  and
      Reiter, Ehud  and
      Sedoc, Joao  and
      Thomson, Craig",
    booktitle = "Proceedings of the Fourth Workshop on Human Evaluation of {NLP} Systems ({H}um{E}val) @ {LREC}-{COLING} 2024",
    month = may,
    year = "2024",
    address = "Torino, Italia",
    publisher = "ELRA and ICCL",
    url = "https://aclanthology.org/2024.humeval-1.1/",
    pages = "1--11"
}

@inproceedings{zerva-etal-2024-findings,
    title = "Findings of the Quality Estimation Shared Task at {WMT} 2024: Are {LLM}s Closing the Gap in {QE}?",
    author = "Zerva, Chrysoula  and
      Blain, Frederic  and
      C. De Souza, Jos{\'e} G.  and
      Kanojia, Diptesh  and
      Deoghare, Sourabh  and
      Guerreiro, Nuno M.  and
      Attanasio, Giuseppe  and
      Rei, Ricardo  and
      Orasan, Constantin  and
      Negri, Matteo  and
      Turchi, Marco  and
      Chatterjee, Rajen  and
      Bhattacharyya, Pushpak  and
      Freitag, Markus  and
      Martins, Andr{\'e}",
    editor = "Haddow, Barry  and
      Kocmi, Tom  and
      Koehn, Philipp  and
      Monz, Christof",
    booktitle = "Proceedings of the Ninth Conference on Machine Translation",
    month = nov,
    year = "2024",
    address = "Miami, Florida, USA",
    publisher = "Association for Computational Linguistics",
    url = "https://aclanthology.org/2024.wmt-1.3/",
    doi = "10.18653/v1/2024.wmt-1.3",
    pages = "82--109"
}

@inproceedings{zeng-etal-2024-towards,
    title = "Towards Multiple References Era {--} Addressing Data Leakage and Limited Reference Diversity in Machine Translation Evaluation",
    author = "Zeng, Xianfeng  and
      Liu, Yijin  and
      Meng, Fandong  and
      Zhou, Jie",
    editor = "Ku, Lun-Wei  and
      Martins, Andre  and
      Srikumar, Vivek",
    booktitle = "Findings of the {A}ssociation for {C}omputational {L}inguistics: {ACL} 2024",
    month = aug,
    year = "2024",
    address = "Bangkok, Thailand",
    publisher = "Association for Computational Linguistics",
    url = "https://aclanthology.org/2024.findings-acl.710/",
    doi = "10.18653/v1/2024.findings-acl.710",
    pages = "11939--11951"
}

@inproceedings{wan-etal-2022-unite,
    title = "{U}ni{TE}: Unified Translation Evaluation",
    author = "Wan, Yu  and
      Liu, Dayiheng  and
      Yang, Baosong  and
      Zhang, Haibo  and
      Chen, Boxing  and
      Wong, Derek  and
      Chao, Lidia",
    editor = "Muresan, Smaranda  and
      Nakov, Preslav  and
      Villavicencio, Aline",
    booktitle = "Proceedings of the 60th Annual Meeting of the {A}ssociation for {C}omputational {L}inguistics (Volume 1: Long Papers)",
    month = may,
    year = "2022",
    address = "Dublin, Ireland",
    publisher = "Association for Computational Linguistics",
    url = "https://aclanthology.org/2022.acl-long.558/",
    doi = "10.18653/v1/2022.acl-long.558",
    pages = "8117--8127"
}

@inproceedings{toral-sanchez-cartagena-2017-multifaceted,
    title = "A Multifaceted Evaluation of Neural versus Phrase-Based Machine Translation for 9 Language Directions",
    author = "Toral, Antonio  and
      S{\'a}nchez-Cartagena, V{\'i}ctor M.",
    editor = "Lapata, Mirella  and
      Blunsom, Phil  and
      Koller, Alexander",
    booktitle = "Proceedings of the 15th Conference of the {E}uropean Chapter of the {A}ssociation for {C}omputational {L}inguistics: Volume 1, Long Papers",
    month = apr,
    year = "2017",
    address = "Valencia, Spain",
    publisher = "Association for Computational Linguistics",
    url = "https://aclanthology.org/E17-1100/",
    pages = "1063--1073"
}

@inproceedings{thompson-post-2020-automatic,
    title = "Automatic Machine Translation Evaluation in Many Languages via Zero-Shot Paraphrasing",
    author = "Thompson, Brian  and
      Post, Matt",
    editor = "Webber, Bonnie  and
      Cohn, Trevor  and
      He, Yulan  and
      Liu, Yang",
    booktitle = "Proceedings of the 2020 Conference on Empirical Methods in Natural Language Processing ({EMNLP})",
    month = nov,
    year = "2020",
    address = "Online",
    publisher = "Association for Computational Linguistics",
    url = "https://aclanthology.org/2020.emnlp-main.8/",
    doi = "10.18653/v1/2020.emnlp-main.8",
    pages = "90--121"
}

@inproceedings{rei-etal-2022-cometkiwi,
    title = "{C}omet{K}iwi: {IST}-Unbabel 2022 Submission for the Quality Estimation Shared Task",
    author = "Rei, Ricardo  and
      Treviso, Marcos  and
      Guerreiro, Nuno M.  and
      Zerva, Chrysoula  and
      Farinha, Ana C  and
      Maroti, Christine  and
      C. de Souza, Jos{\'e} G.  and
      Glushkova, Taisiya  and
      Alves, Duarte  and
      Coheur, Luisa  and
      Lavie, Alon  and
      Martins, Andr{\'e} F. T.",
    editor = {Koehn, Philipp  and
      Barrault, Lo{\"i}c  and
      Bojar, Ond{\v{r}}ej  and
      Bougares, Fethi  and
      Chatterjee, Rajen  and
      Costa-juss{\`a}, Marta R.  and
      Federmann, Christian  and
      Fishel, Mark  and
      Fraser, Alexander  and
      Freitag, Markus  and
      Graham, Yvette  and
      Grundkiewicz, Roman  and
      Guzman, Paco  and
      Haddow, Barry  and
      Huck, Matthias  and
      Jimeno Yepes, Antonio  and
      Kocmi, Tom  and
      Martins, Andr{\'e}  and
      Morishita, Makoto  and
      Monz, Christof  and
      Nagata, Masaaki  and
      Nakazawa, Toshiaki  and
      Negri, Matteo  and
      N{\'e}v{\'e}ol, Aur{\'e}lie  and
      Neves, Mariana  and
      Popel, Martin  and
      Turchi, Marco  and
      Zampieri, Marcos},
    booktitle = "Proceedings of the Seventh Conference on Machine Translation ({WMT})",
    month = dec,
    year = "2022",
    address = "Abu Dhabi, United Arab Emirates (Hybrid)",
    publisher = "Association for Computational Linguistics",
    url = "https://aclanthology.org/2022.wmt-1.60/",
    doi = "10.18653/v1/2022.wmt-1.60",
    pages = "634--645"
}

@inproceedings{rei-etal-2020-comet,
    title = "{COMET}: A Neural Framework for {MT} Evaluation",
    author = "Rei, Ricardo  and
      Stewart, Craig  and
      Farinha, Ana C  and
      Lavie, Alon",
    editor = "Webber, Bonnie  and
      Cohn, Trevor  and
      He, Yulan  and
      Liu, Yang",
    booktitle = "Proceedings of the 2020 Conference on Empirical Methods in Natural Language Processing ({EMNLP})",
    month = nov,
    year = "2020",
    address = "Online",
    publisher = "Association for Computational Linguistics",
    url = "https://aclanthology.org/2020.emnlp-main.213/",
    doi = "10.18653/v1/2020.emnlp-main.213",
    pages = "2685--2702"
}

@inproceedings{rei-etal-2023-scaling,
    title = "Scaling up {C}omet{K}iwi: Unbabel-{IST} 2023 Submission for the Quality Estimation Shared Task",
    author = "Rei, Ricardo  and
      Guerreiro, Nuno M.  and
      Pombal, Jos{\'e}  and
      van Stigt, Daan  and
      Treviso, Marcos  and
      Coheur, Luisa  and
      C. de Souza, Jos{\'e} G.  and
      Martins, Andr{\'e} F. T.",
    editor = "Koehn, Philipp  and
      Haddow, Barry  and
      Kocmi, Tom  and
      Monz, Christof",
    booktitle = "Proceedings of the Eighth Conference on Machine Translation",
    month = dec,
    year = "2023",
    address = "Singapore",
    publisher = "Association for Computational Linguistics",
    url = "https://aclanthology.org/2023.wmt-1.73/",
    doi = "10.18653/v1/2023.wmt-1.73",
    pages = "841--848"
}

@inproceedings{raunak-etal-2021-curious,
    title = "The Curious Case of Hallucinations in Neural Machine Translation",
    author = "Raunak, Vikas  and
      Menezes, Arul  and
      Junczys-Dowmunt, Marcin",
    editor = "Toutanova, Kristina  and
      Rumshisky, Anna  and
      Zettlemoyer, Luke  and
      Hakkani-Tur, Dilek  and
      Beltagy, Iz  and
      Bethard, Steven  and
      Cotterell, Ryan  and
      Chakraborty, Tanmoy  and
      Zhou, Yichao",
    booktitle = "Proceedings of the 2021 Conference of the {N}orth {A}merican Chapter of the {A}ssociation for {C}omputational {L}inguistics: Human Language Technologies",
    month = jun,
    year = "2021",
    address = "Online",
    publisher = "Association for Computational Linguistics",
    url = "https://aclanthology.org/2021.naacl-main.92/",
    doi = "10.18653/v1/2021.naacl-main.92",
    pages = "1172--1183"
}

@inproceedings{qin-specia-2015-truly-exploring,
    title = "Truly Exploring Multiple References for Machine Translation Evaluation",
    author = "Qin, Ying  and
      Specia, Lucia",
    editor = {El-Kahlout, {\.I}lknur Durgar  and
      {\"O}zkan, Mehmed  and
      S{\'a}nchez-Mart{\'i}nez, Felipe  and
      Ram{\'i}rez-S{\'a}nchez, Gema  and
      Hollowood, Fred  and
      Way, Andy},
    booktitle = "Proceedings of the 18th Annual Conference of the {E}uropean Association for Machine Translation",
    month = may,
    year = "2015",
    address = "Antalya, Turkey",
    publisher = "European Association for Machine Translation",
    url = "https://aclanthology.org/W15-4915/",
    pages = "113--120"
}

@inproceedings{proietti-etal-2025-estimating,
    title = "Estimating Machine Translation Difficulty",
    author = "Proietti, Lorenzo  and
      Perrella, Stefano  and
      Zouhar, Vil{\'e}m  and
      Navigli, Roberto  and
      Kocmi, Tom",
    editor = "Christodoulopoulos, Christos  and
      Chakraborty, Tanmoy  and
      Rose, Carolyn  and
      Peng, Violet",
    booktitle = "Findings of the {A}ssociation for {C}omputational {L}inguistics: {EMNLP} 2025",
    month = nov,
    year = "2025",
    address = "Suzhou, China",
    publisher = "Association for Computational Linguistics",
    url = "https://aclanthology.org/2025.findings-emnlp.1317/",
    doi = "10.18653/v1/2025.findings-emnlp.1317",
    pages = "24261--24285",
    ISBN = "979-8-89176-335-7"
}

@inproceedings{post-2018-call,
    title = "A Call for Clarity in Reporting {BLEU} Scores",
    author = "Post, Matt",
    editor = "Bojar, Ond{\v{r}}ej  and
      Chatterjee, Rajen  and
      Federmann, Christian  and
      Fishel, Mark  and
      Graham, Yvette  and
      Haddow, Barry  and
      Huck, Matthias  and
      Yepes, Antonio Jimeno  and
      Koehn, Philipp  and
      Monz, Christof  and
      Negri, Matteo  and
      N{\'e}v{\'e}ol, Aur{\'e}lie  and
      Neves, Mariana  and
      Post, Matt  and
      Specia, Lucia  and
      Turchi, Marco  and
      Verspoor, Karin",
    booktitle = "Proceedings of the Third Conference on Machine Translation: Research Papers",
    month = oct,
    year = "2018",
    address = "Brussels, Belgium",
    publisher = "Association for Computational Linguistics",
    url = "https://aclanthology.org/W18-6319/",
    doi = "10.18653/v1/W18-6319",
    pages = "186--191"
}

@inproceedings{popovic-2015-chrf,
    title = "chr{F}: character n-gram {F}-score for automatic {MT} evaluation",
    author = "Popovi{\'c}, Maja",
    editor = "Bojar, Ond{\v{r}}ej  and
      Chatterjee, Rajan  and
      Federmann, Christian  and
      Haddow, Barry  and
      Hokamp, Chris  and
      Huck, Matthias  and
      Logacheva, Varvara  and
      Pecina, Pavel",
    booktitle = "Proceedings of the Tenth Workshop on Statistical Machine Translation",
    month = sep,
    year = "2015",
    address = "Lisbon, Portugal",
    publisher = "Association for Computational Linguistics",
    url = "https://aclanthology.org/W15-3049/",
    doi = "10.18653/v1/W15-3049",
    pages = "392--395"
}

@inproceedings{noh-etal-2024-beyond,
    title = "Beyond Reference: Evaluating High Quality Translations Better than Human References",
    author = "Noh, Keonwoong  and
      Oh, Seokjin  and
      Jung, Woohwan",
    editor = "Al-Onaizan, Yaser  and
      Bansal, Mohit  and
      Chen, Yun-Nung",
    booktitle = "Proceedings of the 2024 Conference on Empirical Methods in Natural Language Processing",
    month = nov,
    year = "2024",
    address = "Miami, Florida, USA",
    publisher = "Association for Computational Linguistics",
    url = "https://aclanthology.org/2024.emnlp-main.294/",
    doi = "10.18653/v1/2024.emnlp-main.294",
    pages = "5111--5127"
}

@inproceedings{ma-etal-2017-investigation,
    title = "Further Investigation into Reference Bias in Monolingual Evaluation of Machine Translation",
    author = "Ma, Qingsong  and
      Graham, Yvette  and
      Baldwin, Timothy  and
      Liu, Qun",
    editor = "Palmer, Martha  and
      Hwa, Rebecca  and
      Riedel, Sebastian",
    booktitle = "Proceedings of the 2017 Conference on Empirical Methods in Natural Language Processing",
    month = sep,
    year = "2017",
    address = "Copenhagen, Denmark",
    publisher = "Association for Computational Linguistics",
    url = "https://aclanthology.org/D17-1262/",
    doi = "10.18653/v1/D17-1262",
    pages = "2476--2485"
}

@inproceedings{kocmi-etal-2022-findings,
    title = "Findings of the 2022 Conference on Machine Translation ({WMT}22)",
    author = "Kocmi, Tom  and
      Bawden, Rachel  and
      Bojar, Ond{\v{r}}ej  and
      Dvorkovich, Anton  and
      Federmann, Christian  and
      Fishel, Mark  and
      Gowda, Thamme  and
      Graham, Yvette  and
      Grundkiewicz, Roman  and
      Haddow, Barry  and
      Knowles, Rebecca  and
      Koehn, Philipp  and
      Monz, Christof  and
      Morishita, Makoto  and
      Nagata, Masaaki  and
      Nakazawa, Toshiaki  and
      Nov{\'a}k, Michal  and
      Popel, Martin  and
      Popovi{\'c}, Maja",
    editor = {Koehn, Philipp  and
      Barrault, Lo{\"i}c  and
      Bojar, Ond{\v{r}}ej  and
      Bougares, Fethi  and
      Chatterjee, Rajen  and
      Costa-juss{\`a}, Marta R.  and
      Federmann, Christian  and
      Fishel, Mark  and
      Fraser, Alexander  and
      Freitag, Markus  and
      Graham, Yvette  and
      Grundkiewicz, Roman  and
      Guzman, Paco  and
      Haddow, Barry  and
      Huck, Matthias  and
      Jimeno Yepes, Antonio  and
      Kocmi, Tom  and
      Martins, Andr{\'e}  and
      Morishita, Makoto  and
      Monz, Christof  and
      Nagata, Masaaki  and
      Nakazawa, Toshiaki  and
      Negri, Matteo  and
      N{\'e}v{\'e}ol, Aur{\'e}lie  and
      Neves, Mariana  and
      Popel, Martin  and
      Turchi, Marco  and
      Zampieri, Marcos},
    booktitle = "Proceedings of the Seventh Conference on Machine Translation ({WMT})",
    month = dec,
    year = "2022",
    address = "Abu Dhabi, United Arab Emirates (Hybrid)",
    publisher = "Association for Computational Linguistics",
    url = "https://aclanthology.org/2022.wmt-1.1/",
    doi = "10.18653/v1/2022.wmt-1.1",
    pages = "1--45"
}

@inproceedings{kocmi-etal-2023-findings,
    title = "Findings of the 2023 Conference on Machine Translation ({WMT}23): {LLM}s Are Here but Not Quite There Yet",
    author = "Kocmi, Tom  and
      Avramidis, Eleftherios  and
      Bawden, Rachel  and
      Bojar, Ond{\v{r}}ej  and
      Dvorkovich, Anton  and
      Federmann, Christian  and
      Fishel, Mark  and
      Freitag, Markus  and
      Gowda, Thamme  and
      Grundkiewicz, Roman  and
      Haddow, Barry  and
      Koehn, Philipp  and
      Marie, Benjamin  and
      Monz, Christof  and
      Morishita, Makoto  and
      Murray, Kenton  and
      Nagata, Masaaki  and
      Nakazawa, Toshiaki  and
      Popel, Martin  and
      Popovi{\'c}, Maja  and
      Shmatova, Mariya  and
      Suzuki, Jun",
    editor = "Koehn, Philipp  and
      Haddow, Barry  and
      Kocmi, Tom  and
      Monz, Christof",
    booktitle = "Proceedings of the Eighth Conference on Machine Translation",
    month = dec,
    year = "2023",
    address = "Singapore",
    publisher = "Association for Computational Linguistics",
    url = "https://aclanthology.org/2023.wmt-1.1/",
    doi = "10.18653/v1/2023.wmt-1.1",
    pages = "1--42"
}

@inproceedings{kocmi-etal-2024-findings,
    title = "Findings of the {WMT}24 General Machine Translation Shared Task: The {LLM} Era Is Here but {MT} Is Not Solved Yet",
    author = "Kocmi, Tom  and
      Avramidis, Eleftherios  and
      Bawden, Rachel  and
      Bojar, Ond{\v{r}}ej  and
      Dvorkovich, Anton  and
      Federmann, Christian  and
      Fishel, Mark  and
      Freitag, Markus  and
      Gowda, Thamme  and
      Grundkiewicz, Roman  and
      Haddow, Barry  and
      Karpinska, Marzena  and
      Koehn, Philipp  and
      Marie, Benjamin  and
      Monz, Christof  and
      Murray, Kenton  and
      Nagata, Masaaki  and
      Popel, Martin  and
      Popovi{\'c}, Maja  and
      Shmatova, Mariya  and
      Steingr{\'i}msson, Steinth{\'o}r  and
      Zouhar, Vil{\'e}m",
    editor = "Haddow, Barry  and
      Kocmi, Tom  and
      Koehn, Philipp  and
      Monz, Christof",
    booktitle = "Proceedings of the Ninth Conference on Machine Translation",
    month = nov,
    year = "2024",
    address = "Miami, Florida, USA",
    publisher = "Association for Computational Linguistics",
    url = "https://aclanthology.org/2024.wmt-1.1/",
    doi = "10.18653/v1/2024.wmt-1.1",
    pages = "1--46"
}

@inproceedings{kocmi-etal-2025-findings,
    title = "Findings of the {WMT}25 General Machine Translation Shared Task: Time to Stop Evaluating on Easy Test Sets",
    author = "Kocmi, Tom  and
      Artemova, Ekaterina  and
      Avramidis, Eleftherios  and
      Bawden, Rachel  and
      Bojar, Ond{\v{r}}ej  and
      Dranch, Konstantin  and
      Dvorkovich, Anton  and
      Dukanov, Sergey  and
      Fishel, Mark  and
      Freitag, Markus  and
      Gowda, Thamme  and
      Grundkiewicz, Roman  and
      Haddow, Barry  and
      Karpinska, Marzena  and
      Koehn, Philipp  and
      Lakougna, Howard  and
      Lundin, Jessica  and
      Monz, Christof  and
      Murray, Kenton  and
      Nagata, Masaaki  and
      Perrella, Stefano  and
      Proietti, Lorenzo  and
      Popel, Martin  and
      Popovi{\'c}, Maja  and
      Riley, Parker  and
      Shmatova, Mariya  and
      Steingr{\'i}msson, Steinth{\'o}r  and
      Yankovskaya, Lisa  and
      Zouhar, Vil{\'e}m",
    editor = "Haddow, Barry  and
      Kocmi, Tom  and
      Koehn, Philipp  and
      Monz, Christof",
    booktitle = "Proceedings of the Tenth Conference on Machine Translation",
    month = nov,
    year = "2025",
    address = "Suzhou, China",
    publisher = "Association for Computational Linguistics",
    url = "https://aclanthology.org/2025.wmt-1.22/",
    doi = "10.18653/v1/2025.wmt-1.22",
    pages = "355--413",
    ISBN = "979-8-89176-341-8"
}

@inproceedings{juraska-etal-2024-metricx,
    title = "{M}etric{X}-24: The {G}oogle Submission to the {WMT} 2024 Metrics Shared Task",
    author = "Juraska, Juraj  and
      Deutsch, Daniel  and
      Finkelstein, Mara  and
      Freitag, Markus",
    editor = "Haddow, Barry  and
      Kocmi, Tom  and
      Koehn, Philipp  and
      Monz, Christof",
    booktitle = "Proceedings of the Ninth Conference on Machine Translation",
    month = nov,
    year = "2024",
    address = "Miami, Florida, USA",
    publisher = "Association for Computational Linguistics",
    url = "https://aclanthology.org/2024.wmt-1.35/",
    doi = "10.18653/v1/2024.wmt-1.35",
    pages = "492--504"
}

@inproceedings{huang-etal-2024-lost,
    title = "Lost in the Source Language: How Large Language Models Evaluate the Quality of Machine Translation",
    author = "Huang, Xu  and
      Zhang, Zhirui  and
      Geng, Xiang  and
      Du, Yichao  and
      Chen, Jiajun  and
      Huang, Shujian",
    editor = "Ku, Lun-Wei  and
      Martins, Andre  and
      Srikumar, Vivek",
    booktitle = "Findings of the {A}ssociation for {C}omputational {L}inguistics: {ACL} 2024",
    month = aug,
    year = "2024",
    address = "Bangkok, Thailand",
    publisher = "Association for Computational Linguistics",
    url = "https://aclanthology.org/2024.findings-acl.211/",
    doi = "10.18653/v1/2024.findings-acl.211",
    pages = "3546--3562"
}

@article{guerreiro-etal-2024-xcomet,
    title = "x{COMET}: Transparent Machine Translation Evaluation through Fine-grained Error Detection",
    author = "Guerreiro, Nuno M.  and
      Rei, Ricardo  and
      van Stigt, Daan  and
      Coheur, Luisa  and
      Colombo, Pierre  and
      Martins, Andr{\'e} F. T.",
    journal = "Transactions of the Association for Computational Linguistics",
    volume = "12",
    year = "2024",
    address = "Cambridge, MA",
    publisher = "MIT Press",
    url = "https://aclanthology.org/2024.tacl-1.54/",
    doi = "10.1162/tacl_a_00683",
    pages = "979--995"
}

@article{guerreiro-etal-2023-hallucinations,
    title = "Hallucinations in Large Multilingual Translation Models",
    author = "Guerreiro, Nuno M.  and
      Alves, Duarte M.  and
      Waldendorf, Jonas  and
      Haddow, Barry  and
      Birch, Alexandra  and
      Colombo, Pierre  and
      Martins, Andr{\'e} F. T.",
    journal = "Transactions of the Association for Computational Linguistics",
    volume = "11",
    year = "2023",
    address = "Cambridge, MA",
    publisher = "MIT Press",
    url = "https://aclanthology.org/2023.tacl-1.85/",
    doi = "10.1162/tacl_a_00615",
    pages = "1500--1517"
}

@inproceedings{freitag-etal-2021-results,
    title = "Results of the {WMT}21 Metrics Shared Task: Evaluating Metrics with Expert-based Human Evaluations on {TED} and News Domain",
    author = "Freitag, Markus  and
      Rei, Ricardo  and
      Mathur, Nitika  and
      Lo, Chi-kiu  and
      Stewart, Craig  and
      Foster, George  and
      Lavie, Alon  and
      Bojar, Ond{\v{r}}ej",
    editor = "Barrault, Loic  and
      Bojar, Ondrej  and
      Bougares, Fethi  and
      Chatterjee, Rajen  and
      Costa-jussa, Marta R.  and
      Federmann, Christian  and
      Fishel, Mark  and
      Fraser, Alexander  and
      Freitag, Markus  and
      Graham, Yvette  and
      Grundkiewicz, Roman  and
      Guzman, Paco  and
      Haddow, Barry  and
      Huck, Matthias  and
      Yepes, Antonio Jimeno  and
      Koehn, Philipp  and
      Kocmi, Tom  and
      Martins, Andre  and
      Morishita, Makoto  and
      Monz, Christof",
    booktitle = "Proceedings of the Sixth Conference on Machine Translation",
    month = nov,
    year = "2021",
    address = "Online",
    publisher = "Association for Computational Linguistics",
    url = "https://aclanthology.org/2021.wmt-1.73/",
    pages = "733--774"
}

@inproceedings{freitag-etal-2020-bleu,
    title = "{BLEU} might be Guilty but References are not Innocent",
    author = "Freitag, Markus  and
      Grangier, David  and
      Caswell, Isaac",
    editor = "Webber, Bonnie  and
      Cohn, Trevor  and
      He, Yulan  and
      Liu, Yang",
    booktitle = "Proceedings of the 2020 Conference on Empirical Methods in Natural Language Processing ({EMNLP})",
    month = nov,
    year = "2020",
    address = "Online",
    publisher = "Association for Computational Linguistics",
    url = "https://aclanthology.org/2020.emnlp-main.5/",
    doi = "10.18653/v1/2020.emnlp-main.5",
    pages = "61--71"
}

@article{freitag-etal-2021-experts,
    title = "Experts, Errors, and Context: A Large-Scale Study of Human Evaluation for Machine Translation",
    author = "Freitag, Markus  and
      Foster, George  and
      Grangier, David  and
      Ratnakar, Viresh  and
      Tan, Qijun  and
      Macherey, Wolfgang",
    editor = "Roark, Brian  and
      Nenkova, Ani",
    journal = "Transactions of the Association for Computational Linguistics",
    volume = "9",
    year = "2021",
    address = "Cambridge, MA",
    publisher = "MIT Press",
    url = "https://aclanthology.org/2021.tacl-1.87/",
    doi = "10.1162/tacl_a_00437",
    pages = "1460--1474"
}

@inproceedings{freitag-etal-2020-human,
    title = "Human-Paraphrased References Improve Neural Machine Translation",
    author = "Freitag, Markus  and
      Foster, George  and
      Grangier, David  and
      Cherry, Colin",
    editor = {Barrault, Lo{\"i}c  and
      Bojar, Ond{\v{r}}ej  and
      Bougares, Fethi  and
      Chatterjee, Rajen  and
      Costa-juss{\`a}, Marta R.  and
      Federmann, Christian  and
      Fishel, Mark  and
      Fraser, Alexander  and
      Graham, Yvette  and
      Guzman, Paco  and
      Haddow, Barry  and
      Huck, Matthias  and
      Yepes, Antonio Jimeno  and
      Koehn, Philipp  and
      Martins, Andr{\'e}  and
      Morishita, Makoto  and
      Monz, Christof  and
      Nagata, Masaaki  and
      Nakazawa, Toshiaki  and
      Negri, Matteo},
    booktitle = "Proceedings of the Fifth Conference on Machine Translation",
    month = nov,
    year = "2020",
    address = "Online",
    publisher = "Association for Computational Linguistics",
    url = "https://aclanthology.org/2020.wmt-1.140/",
    doi = "10.18653/v1/2020.wmt-1.140",
    pages = "1183--1192"
}

@inproceedings{fomicheva-specia-2016-reference,
    title = "Reference Bias in Monolingual Machine Translation Evaluation",
    author = "Fomicheva, Marina  and
      Specia, Lucia",
    editor = "Erk, Katrin  and
      Smith, Noah A.",
    booktitle = "Proceedings of the 54th Annual Meeting of the {A}ssociation for {C}omputational {L}inguistics (Volume 2: Short Papers)",
    month = aug,
    year = "2016",
    address = "Berlin, Germany",
    publisher = "Association for Computational Linguistics",
    url = "https://aclanthology.org/P16-2013/",
    doi = "10.18653/v1/P16-2013",
    pages = "77--82"
}

@inproceedings{deutsch-etal-2025-wmt24,
    title = "{WMT}24++: Expanding the Language Coverage of {WMT}24 to 55 Languages {\&} Dialects",
    author = "Deutsch, Daniel  and
      Briakou, Eleftheria  and
      Caswell, Isaac Rayburn  and
      Finkelstein, Mara  and
      Galor, Rebecca  and
      Juraska, Juraj  and
      Kovacs, Geza  and
      Lui, Alison  and
      Rei, Ricardo  and
      Riesa, Jason  and
      Rijhwani, Shruti  and
      Riley, Parker  and
      Salesky, Elizabeth  and
      Trabelsi, Firas  and
      Winkler, Stephanie  and
      Zhang, Biao  and
      Freitag, Markus",
    editor = "Che, Wanxiang  and
      Nabende, Joyce  and
      Shutova, Ekaterina  and
      Pilehvar, Mohammad Taher",
    booktitle = "Findings of the {A}ssociation for {C}omputational {L}inguistics: {ACL} 2025",
    month = jul,
    year = "2025",
    address = "Vienna, Austria",
    publisher = "Association for Computational Linguistics",
    url = "https://aclanthology.org/2025.findings-acl.634/",
    doi = "10.18653/v1/2025.findings-acl.634",
    pages = "12257--12284",
    ISBN = "979-8-89176-256-5"
}

@inproceedings{agrawal-etal-2021-assessing,
    title = "Assessing Reference-Free Peer Evaluation for Machine Translation",
    author = "Agrawal, Sweta  and
      Foster, George  and
      Freitag, Markus  and
      Cherry, Colin",
    editor = "Toutanova, Kristina  and
      Rumshisky, Anna  and
      Zettlemoyer, Luke  and
      Hakkani-Tur, Dilek  and
      Beltagy, Iz  and
      Bethard, Steven  and
      Cotterell, Ryan  and
      Chakraborty, Tanmoy  and
      Zhou, Yichao",
    booktitle = "Proceedings of the 2021 Conference of the {N}orth {A}merican Chapter of the {A}ssociation for {C}omputational {L}inguistics: Human Language Technologies",
    month = jun,
    year = "2021",
    address = "Online",
    publisher = "Association for Computational Linguistics",
    url = "https://aclanthology.org/2021.naacl-main.91/",
    doi = "10.18653/v1/2021.naacl-main.91",
    pages = "1158--1171"
}

@misc{qwen3.5,
    title  = {{Qwen3.5}: Towards Native Multimodal Agents},
    author = {{Qwen Team}},
    month  = {February},
    year   = {2026},
    url    = {https://qwen.ai/blog?id=qwen3.5}
}

@inproceedings{Moosa2024MTRankerRM,
    title={MT-Ranker: Reference-free machine translation evaluation by inter-system ranking},
    author={Ibraheem Muhammad Moosa and Rui Zhang and Wenpeng Yin},
    booktitle={The Twelfth International Conference on Learning Representations},
    year={2024},
    url={https://openreview.net/forum?id=Rry1SeSOQL}
}

@inproceedings{zouhar-etal-2026-generating,
    title = "Generating Difficult-to-Translate Texts",
    author = "Zouhar, Vil{\'e}m  and
      Xu, Wenda  and
      Riley, Parker  and
      Juraska, Juraj  and
      Finkelstein, Mara  and
      Freitag, Markus  and
      Deutsch, Daniel",
    editor = "Chen, Pinzhen  and
      Zouhar, Vil{\'e}m  and
      Hu, Hanxu  and
      Khanuja, Simran  and
      Zhu, Wenhao  and
      Haddow, Barry  and
      Birch, Alexandra  and
      Aji, Alham Fikri  and
      Sennrich, Rico  and
      Hooker, Sara",
    booktitle = "Proceedings of the First Workshop on Multilingual Multicultural Evaluation",
    month = mar,
    year = "2026",
    address = "Rabat, Morocco",
    publisher = "Association for Computational Linguistics",
    url = "https://aclanthology.org/2026.mme-main.14/",
    doi = "10.18653/v1/2026.mme-main.14",
    pages = "204--219",
    ISBN = "979-8-89176-368-5"
}

@inproceedings{zhang-2020-BERTScore,
    title={BERTScore: Evaluating Text Generation with BERT},
    author={Tianyi Zhang and Varsha Kishore and Felix Wu and Kilian Q. Weinberger and Yoav Artzi},
    booktitle={International Conference on Learning Representations},
    year={2020},
    url={https://openreview.net/forum?id=SkeHuCVFDr}
}

\begin{table*}[t]
\centering\small
\begin{tabular}{llrrrl}
\toprule
Year & Pair & Systems & Units & Raters/seg & Typology \\
\midrule
2022 & En--De & 16 & 21{,}040 & 1 & Hierarchical \\
2022 & En--Ru & 16 & 19{,}440 & 1 & Flat, entity and hallucination branches \\
2022 & En--Zh & 13 & 26{,}481 & 3 & Hierarchical \\
2022 & Zh--En & 16 & 30{,}000 & 1 & Hierarchical, no reference annotated \\
2023 & En--De & 13 & 5{,}980 & 1 & Hierarchical \\
2023 & He--En & 13 & 10{,}660 & 1 & Flat, entity branch \\
2023 & Zh--En & 16 & 18{,}832 & 1 & Hierarchical \\
2024 & En--De & 19 & 9{,}253 & 1 & Hierarchical \\
2024 & En--Es & 15 & 9{,}345 & 1 & Flat, entity branch \\
2024 & Ja--Zh & 15 & 8{,}400 & 1 & Hierarchical \\
\bottomrule
\end{tabular}
\caption{WMT general MT MQM releases used in this study. Units counts distinct system and segment pairs after annotator collapsing. Typology names the label set before normalisation, and marks the three releases that separate named entity errors from mistranslation.}
\label{tab:releases}
\end{table*}

\begin{table*}[t]
\centering\small
\begin{tabular}{llrrrrlrr}
\toprule
Year & Pair & Ref & MQM & Err/seg & \%Seg & Best system & Best MQM & Rank \\
\midrule
2022 & En--De & refA & 0.64 & 0.44 & 30.3 & Online-W & 0.79 & 1 / 16 \\
2022 & En--De & refB & 0.91 & 0.56 & 35.6 & Online-W & 0.79 & 3 / 16 \\
2022 & En--Ru & refA & 1.22 & 0.66 & 40.5 & Online-W & 1.47 & 1 / 16 \\
2022 & En--Zh & refB & 1.42 & 0.90 & 86.2 & Lan-Bridge & 1.57 & 1 / 13 \\
2023 & En--De & refA & 3.09 & 1.80 & 58.9 & GPT4-5shot & 3.81 & 1 / 13 \\
2023 & He--En & refA & 1.18 & 0.46 & 30.2 & GPT4-5shot & 1.33 & 1 / 13 \\
2023 & Zh--En & refA & 4.86 & 2.77 & 86.7 & Lan-BridgeMT & 2.10 & 12 / 16 \\
2024 & En--De & refB & 1.96 & 1.18 & 41.9 & Dubformer & 1.69 & 6 / 19 \\
2024 & En--De & refA & 2.23 & 1.25 & 47.6 & Dubformer & 1.69 & 9 / 19 \\
2024 & En--Es & refA & 0.55 & 0.33 & 18.6 & GPT-4 & 0.12 & 8 / 15 \\
2024 & Ja--Zh & refA & 1.38 & 0.58 & 31.8 & Claude-3.5 & 1.25 & 2 / 15 \\
\bottomrule
\end{tabular}
\caption{Human references in the WMT MQM campaigns. Lower MQM is better. \%Seg is the share of reference segments carrying at least one error. Rank places the reference among all annotated systems in its own campaign. Scores are comparable within a campaign but not across campaigns.}
\label{tab:refprofile}
\end{table*}

\newpage
\appendix
\crefalias{section}{appendix}
\crefalias{subsection}{appendix}
\crefalias{subsubsection}{appendix}

\section*{Appendix}

\section{Reference Errors in WMT MQM}
\label{appx:mqm}

\subsection{Data and Procedure}
\label{appx:mqm-procedure}

WMT scores the human reference as one system among the MT submissions in its MQM campaigns, which makes reference quality directly measurable rather than assumed. We collect the general MT campaigns from 2022 through 2024. The releases differ in schema, in error typology and in annotation density, and \Cref{tab:releases} summarises what each contributes.

Three normalization steps are needed before the releases can be compared. First, the typologies differ. The 2022 En$\rightarrow$Ru release uses a flat label set with underscores, the 2023 He$\rightarrow$En and 2024 En$\rightarrow$Es releases drop the branch prefix, and the remaining releases follow the hierarchy of \citet{freitag-etal-2021-experts}. We fold all of them onto that hierarchy, which also recovers two labels that no other release separates, namely an explicit named entity branch and an explicit hallucination branch. Second, hands-on-the-wheel items are removed. Those rows record whether an annotator caught an error that the organizers injected deliberately, so they describe annotator vigilance rather than translation quality, and counting them would place synthetic corruptions inside a tally of naturally occurring ones. Third, source-side errors are excluded because they do not reflect target quality.

Segment scores follow the MQM weighting of \citet{freitag-etal-2021-experts}, with 25 points for non-translation, 5 for a major error, 1 for a minor error and 0.1 for minor punctuation. Critical severity is scored as Major. System ranking averages over annotators before scoring, which matters only for the 2022 En$\rightarrow$Zh release, the one release annotated three times per segment. Error counts and category shares are reported per language pair rather than pooled, since pooling would let that release and its 26{,}481 annotation units dominate every aggregate.

\begin{table}[t]
\centering\small
\setlength{\tabcolsep}{4pt}
\begin{tabular}{llcccc}
\toprule
& & \multicolumn{2}{c}{Strict} & \multicolumn{2}{c}{Loose} \\
\cmidrule(lr){3-4}\cmidrule(lr){5-6}
Year & Pair & All & Major & All & Major \\
\midrule
2022 & En--De & 19.7 & 40.3 & 27.1 & 63.7 \\
2022 & En--Ru & 29.8 & 17.1 & 37.1 & 25.9 \\
2022 & En--Zh & 46.6 & 29.7 & 54.6 & 46.8 \\
2023 & En--De & 28.0 & 31.6 & 39.3 & 49.1 \\
2023 & He--En & 37.6 & 32.2 & 39.5 & 36.2 \\
2023 & Zh--En & 30.6 & 39.1 & 32.2 & 44.5 \\
2024 & En--De & 38.2 & 41.7 & 47.9 & 62.5 \\
2024 & En--Es & 61.8 & 50.0 & 74.0 & 76.2 \\
2024 & Ja--Zh & 47.7 & 58.0 & 47.7 & 58.0 \\
\midrule
& Median & 37.6 & 39.1 & 39.5 & 49.1 \\
\bottomrule
\end{tabular}
\caption{Share of reference error annotations corresponding to a \mirrorbench contamination type. Strict counts MQM category matches only. Loose additionally counts span-heuristic attributions, whose precision is reported in \Cref{tab:ruleprec}. Zh--En 2022 is omitted because its release carries no reference annotation.}
\label{tab:coverage}
\end{table}

\subsection{Reference Quality}
\label{appx:mqm-quality}

\Cref{tab:refprofile} gives the full ranking. The reference ranks first in five of the six 2022 and 2023 pairs and in none of the 2024 pairs. The best system in 2024 outscores the reference in all three pairs, by 0.43 MQM points in En$\rightarrow$Es, 0.27 in En$\rightarrow$De and 0.13 in Ja$\rightarrow$Zh. Absolute scores are not comparable across campaigns because raters, guidelines and domains differ, so the ranking column carries the comparison and the score columns are given for reference only.

The share of reference segments carrying at least one annotated error ranges from 18.6\% in 2024 En$\rightarrow$Es to 86.7\% in 2023 Zh$\rightarrow$En. That spread tracks annotation practice as much as reference quality, since the two pairs at the top of the range are also the two with the highest error counts on their MT submissions.

\subsection{Coverage of the Contamination Types}
\label{appx:mqm-coverage}

\Cref{tab:coverage} maps every reference error annotation onto the seven contamination types. The strict columns count only annotations whose MQM category corresponds to a type. The loose columns additionally count annotations attributed by the span heuristic described in \Cref{appx:entity}, and we report them to show how much of an apparent coverage figure that heuristic supplies.

Coverage does not concentrate on severe errors. Under the strict definition the major figure exceeds the overall figure in five of nine pairs, and the two medians differ by 1.5 points. The uncovered mass splits in two. Style and fluency errors supply its largest share and almost never reach major severity, with a single major error among the several thousand style errors in the 2022 En$\rightarrow$De reference set. Mistranslation that resolves into neither a named entity nor a polarity change is less frequent and often major, and it accounts for most of the major reference errors the taxonomy does not probe. This is a consequence of the design rather than an oversight, since a probe that isolates one corruption at a time cannot represent an error category defined by the absence of any narrower structure.

\begin{tcolorbox}[
  width=\columnwidth,
  enlarge left by=0pt,
  left=6pt, right=6pt, top=6pt, bottom=6pt,
  colback=gray!5,
  colframe=gray!40,
  boxrule=0.5pt,
  arc=2pt,
  before skip=6pt, after skip=6pt,
]
\setlength{\parindent}{0pt}
\paragraph{\protect\circled{tPink!80}{A} \protect\MinorNoiseCat.} The most frequent
type in the reference sets at 142.9 occurrences per thousand segments and the least
severe, major in 1.1\% of cases. Retained as an upper bound on metric sensitivity
rather than as a claim about frequency.

\paragraph{\protect\circled{tPink!80}{A} \protect\SourceCopyCat.} Annotated as
Accuracy/Source language fragment in 7 of 9 pairs, at 20.2 per thousand segments
and major in 25.3\% of cases.

\paragraph{\protect\circled{tGreen!70}{B} \protect\WrongLangCat.} Attributed in 4
pairs, all of which mix scripts between source and target, by a span rule we do not
validate. The rule may be recovering untranslated fragments rather than reference
segments written in a third language.

\paragraph{\protect\circled{tGreen!70}{B} \protect\UnalignedCat.} The rarest type,
at 2.1 per thousand segments in 7 pairs, and the most severe when it occurs, major
in 65.0\% of cases.

\paragraph{\protect\circled{tBlue!70}{C} \protect\HalluCat.} Annotated as
Accuracy/Addition in every pair, and as an explicit hallucination category in the
2022 En$\rightarrow$Ru release, at 21.7 per thousand segments and major in 51.5\%
of cases.

\paragraph{\protect\circled{tBlue!70}{C} \protect\PolarityCat.} The largest type
identified by a span rule, at 56.1 per thousand segments and major in 48.4\% of
cases. Its rule compares negation cue counts between source and target, which is
sensitive to how each language realises negation, so we report the figure without
building on it.

\paragraph{\protect\circled{tBlue!70}{C} \protect\ArgSwapCat.} Annotated at 2.6\%, 7.4\% and 14.7\% of reference errors in the three releases that separate named entities from mistranslation, Major in 19\% of those cases at the lowest, and confined to a single entity in 87\% of affected segments.
\end{tcolorbox}

\begin{table}[t]
\centering\small
\setlength{\tabcolsep}{4pt}
\begin{tabular}{llrrrr}
\toprule
Type & Src & Pairs & Per 1k & Err\% & Maj\% \\
\midrule
\Cat{Minor Noise}      & cat & 9 & 142.9 & 14.5 & 1.1 \\
\Cat{Content Addition} & cat & 9 & 21.7 & 3.2 & 51.5 \\
\Cat{Source Copy}      & cat & 7 & 20.2 & 2.0 & 25.3 \\
\Cat{Entity Swap}      & cat & 3 & 24.1 & 7.4 & 44.0 \\
\midrule
\Cat{Polarity Flip}    & rule & 9 & 56.1 & 6.3 & 48.4 \\
\Cat{Entity Swap}      & rule & 8 & 13.9 & 1.1 & 52.0 \\
\Cat{Wrong Language}   & rule & 4 & 10.6 & 1.5 & 23.6 \\
\Cat{Unaligned Sent}   & rule & 7 & 2.1 & 0.3 & 65.0 \\
\bottomrule
\end{tabular}
\caption{Observed rate of each contamination type in the reference sets. Src marks
whether the type was identified by an MQM category or by a span rule.
\protect\Cat{Entity Swap} appears twice because only three releases annotate named
entities as a category, so the two rows cover disjoint pairs. Pairs counts the
language pairs in which the type occurs. Per 1k and Err\% are medians across those
pairs, Maj\% pools the annotations.}
\label{tab:typerates}
\end{table}

\subsection{Observed Rate of Each Contamination Type}
\label{appx:mqm-types}
\Cref{tab:typerates} gives these rates together. The Src column separates the types an MQM category names directly from those only a span rule identifies, and the second group inherits the reliability discussed in \Cref{appx:entity}. \Cat{Entity Swap} appears twice because three releases annotate named entities as a category and the rest do not, and the two rows cover disjoint sets of pairs.

% ---------------------------------------------------------------------------
\begin{table}[t]
\centering\small
\setlength{\tabcolsep}{4pt}
\begin{tabular}{lrrr}
\toprule
Pair & \%Ref errors & \%Sys errors & \%Major \\
\midrule
2022 En--Ru & 2.6 & 4.0 & 19.0 \\
2024 En--Es & 7.4 & 1.4 & 100.0 \\
2023 He--En & 14.7 & 16.5 & 38.2 \\
\bottomrule
\end{tabular}
\caption{Named entity errors in the three releases that annotate them as a separate MQM category. \%Major is the share of the reference-side cases annotated at major severity. En--Es rests on fifteen reference cases.}
\label{tab:entitygold}
\end{table}

\begin{table}[t]
\centering\small
\setlength{\tabcolsep}{4pt}
\resizebox{\columnwidth}{!}{
\begin{tabular}{lrrc}
\toprule
Rule & $n$ & Precision & Per-pair range \\
\midrule
Capitalized token in span & 600 & 0.64 & 0.00--0.65 \\
Source entity in span & 81 & 0.26 & 0.15--0.32 \\
Source entity absent & 3{,}952 & 0.18 & 0.06--0.20 \\
Digit or number word & 173 & 0.15 & 0.04--0.28 \\
\bottomrule
\end{tabular}
}
\caption{Precision of each span rule against the gold named entity category, pooled over the three labelled pairs. The per-pair range shows that precision is a property of the language pair rather than of the rule.}
\label{tab:ruleprec}
\end{table}

\begin{table}[t]
\centering\small
\begin{tabular}{lrrr}
\toprule
Held-out pair & Predicted & Estimated & Gold \\
\midrule
2022 En--Ru & 74 & 19.6 & 21 \\
2024 En--Es & 35 & 8.1 & 15 \\
2023 He--En & 21 & 0.0 & 55 \\
\bottomrule
\end{tabular}
\caption{Leave-one-out test of whether rule precision transfers across language pairs. Each pair is estimated from the precision measured on the other two, on the reference side. Predicted counts raw rule firings before correction.}
\label{tab:loo}
\end{table}

\section{Named Entity Errors and the Limits of Rule-Based Detection}
\label{appx:entity}

\subsection{Rates Where the Category Is Annotated}

Three releases annotate named entity errors as a separate MQM category instead of folding them into mistranslation. \Cref{tab:entitygold} gives the resulting rates. Named entity errors account for 2.6 to 14.7\% of reference errors across the three, and the spread follows the language pair rather than the year. He$\rightarrow$En sits highest on both sides of the comparison, at 14.7\% of reference errors and 16.5\% of system errors. En$\rightarrow$Es runs the other way, where the reference rate of 7.4\% exceeds the 1.4\% measured on system output.

Severity separates these errors from the categories that outnumber them. Major severity covers 19.0\% of the En$\rightarrow$Ru cases, 38.2\% of the He$\rightarrow$En cases and all fifteen En$\rightarrow$Es cases, against one major error among the several thousand style errors in the 2022 En$\rightarrow$De reference set.

\subsection{Why We Do Not Estimate the Remaining Pairs}

The other six pairs do not separate named entities, so a rate for them would have to come from a heuristic. We built one and then measured it, and the measurement is why no such rate appears in this paper.

The heuristic inspects the marked error span and fires on four kinds of evidence, namely a digit or number word in the span, a capitalized source token appearing inside the span, a capitalized source token absent from the target, and a capitalized token inside the span on a target side that carries case. \Cref{tab:ruleprec} gives the precision of each rule on the pairs that have gold labels. Pooled across those pairs the best rule reaches 0.64 and the weakest reaches 0.15, but the per-pair figures matter more than the pooled ones. The same rule that reaches 0.65 on He$\rightarrow$En reaches 0.00 on En$\rightarrow$Es. Precision is a property of the language pair here, not of the rule.

\Cref{tab:loo} tests the transfer directly by estimating each labeled pair from the precision measured on the other two. The procedure recovers 21 of 21 entity errors in En$\rightarrow$Ru, 8 of 15 in En$\rightarrow$Es and none of the 55 in He$\rightarrow$En. The error runs in both directions and is not correctable by a single scaling factor. Applying the same procedure to a pair with no gold labels would therefore produce a number with no measurable relationship to the quantity it names.

We read this as evidence about the error type rather than about our implementation. A detector with access to both the source and the target, and to the annotated span, cannot identify named entity errors at a precision that transfers across language pairs. That is consistent with the central result of \Cref{sec:result}, where no metric detects \Cat{Entity Swap} while the translation model remains robust to it. Named entity errors are locally fluent and require source checking rather than target reading, and surface evidence does not substitute for that check.

% ---------------------------------------------------------------------------

\begin{table*}[t]
\centering
\small
\begin{minipage}[t]{0.40\textwidth}
\centering
\setlength{\tabcolsep}{4pt}
\begin{tabular}{l r r r r r}
\toprule
 & \textbf{\#Seg} & \textbf{En} & \textbf{De} & \textbf{Zh} & \textbf{Ko} \\
\midrule
Social   & 291 & 27.6 & 28.4 & 26.9 & 34.0 \\
Literary & 168 & 53.2 & 52.7 & 51.3 & 67.9 \\
News     & 135 & 65.5 & 63.1 & 61.8 & 82.6 \\
Speech   & 111 & 84.9 & 80.7 & 77.4 & 95.5 \\
\midrule
Total    & 705 & 47.4 & 46.6 & 45.0 & 58.7 \\
\midrule
Ratio    & --  & 1.00 & 0.98 & 0.95 & 1.24 \\
\bottomrule
\end{tabular}
\subcaption{\textbf{Corpus composition.} Per-domain segment counts and average tokens per segment by language. The bottom row reports the macro-averaged length ratio against the English source.}
\label{tab:data-stats}
\end{minipage}%
\hfill
\begin{minipage}[t]{0.56\textwidth}
\centering
\setlength{\tabcolsep}{4pt}
\begin{tabular}{l r r r r r r}
\toprule
 & \multicolumn{3}{c}{\textbf{Length div.\ (\%)}} & \multicolumn{3}{c}{\textbf{TTR}} \\
\cmidrule(lr){2-4} \cmidrule(lr){5-7}
\textbf{Perturbation} & \textbf{De} & \textbf{Ko} & \textbf{Zh} & \textbf{De} & \textbf{Ko} & \textbf{Zh} \\
\midrule
\textit{Clean}   & --    & --    & --     & 0.244 & 0.145 & 0.217 \\
\midrule
\MinorNoiseCat   & +10.0 & +17.6 & +19.9  & 0.423 & 0.165 & 0.376 \\
\SourceCopyCat   &  +3.7 & -18.9 & -79.3  & 0.237 & 0.204 & 0.391 \\
\WrongLangCat    & -57.1 &  +1.8 & +65.7  & 0.525 & 0.191 & 0.112 \\
\UnalignedCat    & +24.2 & +24.4 & +27.6  & 0.233 & 0.142 & 0.209 \\
\HalluCat        & +62.2 & +62.6 & +112.7 & 0.154 & 0.094 & 0.114 \\
\PolarityCat     &  +4.6 &  +4.7 &  +6.6  & 0.231 & 0.141 & 0.207 \\
\ArgSwapCat     &  +0.3 &  -1.4 &  -0.5  & 0.272 & 0.176 & 0.267 \\
\bottomrule
\end{tabular}
\subcaption{\textbf{Surface-level signatures of each perturbation.} \textbf{Length div.} is the percentage change in token count of the contaminated reference relative to the clean reference. \textbf{TTR} = type-token ratio computed with language-specific tokenizers. All contaminations use 705 segments per language, except \Cat{Entity Swap}, which is restricted to segments where spaCy detects at least one named entity (425/De, 366/Ko, 347/Zh).}
\label{tab:perturb-stats}
\end{minipage}
\caption{\mirrorbench statistics.}
\label{tab:stats-combined}
\end{table*}

\section{Benchmark Construction Details}
\label{appx:construction}

\subsection{Data Construction}

We exclude segments shorter than 10 words so that perturbations have enough surface to operate on, and we discard additional segments to balance contamination levels across domains. Detailed statistics are given in \Cref{tab:stats-combined}.

\subsection{Data Statistics}
\label{appx:stats}

\Cref{tab:stats-combined} reports the statistics of \mirrorbench. Counts are identical across the three language pairs. Token counts use the SacreBLEU 13a tokenizer~\citep{post-2018-call} for English and German, Jieba\footnote{\url{https://github.com/fxsjy/jieba}} for Chinese, and MeCab-Ko\footnote{\url{https://bitbucket.org/eunjeon/mecab-ko}} for Korean.

The left panel shows segment length growing steadily from \textit{social} to \textit{speech}, with the longest domain averaging roughly three times the tokens of the shortest. Reference-to-source length ratios follow expected typological patterns. Korean produces 24\% more tokens than the English source due to its agglutinative morphology, Chinese produces 5\% fewer due to its logographic script, and German tracks English almost exactly.

The right panel quantifies the unintended surface effects of each perturbation. \Cat{Content Addition} lengthens the reference by 62 to 113\% by appending an adjacent segment. \Cat{Wrong Language} and \Cat{Source Copy} show language-dependent length changes because token granularity differs across the source, target, and substitution languages. \Cat{Minor Noise} sharply raises TTR in German and Chinese (from 0.244 to 0.423, and 0.217 to 0.376) because character-level reordering produces non-word surface forms that count as new types. By contrast, \Cat{Polarity Flip} and \Cat{Entity Swap} keep length divergence within $\pm 7\%$ and TTR within 0.03 of the clean reference across all three languages, since both perturbations rewrite the reference at the semantic level while preserving its lexical and structural shape. These two are therefore the perturbations least detectable from surface features, a property we revisit in \Cref{sec:result}.

\begin{table*}[t]
\centering
\small
\renewcommand{\arraystretch}{1.3}
\setlength{\tabcolsep}{6pt}
\begin{tabular}{@{}p{0.04\linewidth} p{0.16\linewidth} p{0.72\linewidth}@{}}
\toprule
\textbf{Cat.} & \textbf{Type} & \textbf{Text} \\
\midrule
\multicolumn{2}{@{}l}{\textit{Source (En)}}
& On a full charge, it gives me a runtime of about 4 hours. \\
\multicolumn{2}{@{}l}{\textit{Clean reference (De)}}
& Wenn sie aufgeladen ist, habe ich eine Laufzeit von etwa 4 Stunden. \\
\midrule
\protect\circled{tPink!80}{A}  & Minor Noise
& Wenn sie aufgeladen \hl{,sti} \hl{beah} ich eine \hl{aeuLitzf} \hl{ovn} \hl{weta} 4 \hl{dntneu.S} \\
\addlinespace[2pt]
\protect\circled{tPink!80}{A} & Source Copy
& \hl{runtime} sie aufgeladen \hl{charge,} \hl{hours.} ich eine \hl{full} \hl{full} etwa \hl{runtime} \hl{charge,} \\
\protect\circled{tGreen!70}{B} & Wrong Language
& \hl{\begin{CJK}{UTF8}{gbsn}充满电后，可以运行大约4个小时。\end{CJK}} \\
\protect\circled{tGreen!70}{B} & Unaligned Sent
& \hl{Das} \hl{nenne} \hl{ich} \hl{mal} \hl{ein} \hl{Feature.} \hl{So} \hl{bekommen} \hl{die} \hl{V\"ogel} \hl{auch} \hl{etwas} \hl{Privatsph\"are.} \\
\addlinespace[2pt]
\protect\circled{tBlue!70}{C}  & Content Addition
& Wenn sie aufgeladen ist, habe ich eine Laufzeit von etwa 4 Stunden. \hl{Das} \hl{nenne} \hl{ich} \hl{mal} \hl{ein} \hl{Feature.} \hl{So} \hl{bekommen} \\
\rowcolor{red!6}
\protect\circled{tBlue!70}{C}  & Polarity Flip
& Wenn sie aufgeladen ist, habe ich \textbf{\hl{keine}} Laufzeit von etwa 4 Stunden. \\
\rowcolor{red!6}
\protect\circled{tBlue!70}{C}  & Entity Swap
& \textbf{\hl{Weber}} aufgeladen h\"alt es ungef\"ahr 4 Stunden. \\
\bottomrule
\end{tabular}
\caption{Examples of the seven perturbation types in \mirrorbench, applied to a single English source and its German reference. Modified or inserted spans are \hl{highlighted}, with the word that flips meaning shown in \textbf{bold}. Categories follow \cref{subsec:perturbations} (\protect\circled{tPink!80}{A} Surface noise, \protect\circled{tGreen!70}{B} Irrelevant content, \protect\circled{tBlue!70}{C} Hallucination). The two shaded rows mark the categories that most consistently subvert metrics in our experiments.}
\label{tab:real_ex}
\end{table*}

\subsection{Additional Benchmark Examples}
\label{appx:examples}

To complement the single-segment example in \Cref{tab:bias_type}, we provide additional samples drawn directly from \mirrorbench. The samples below come from different source segments in En-De, and illustrate how each perturbation type appears in realistic content. We include one example per perturbation type, with the English source, the clean German reference, and the contaminated reference shown side by side. For the multi-level perturbations (\Cat{Minor Noise}, \Cat{Source Copy}, \Cat{Entity Swap}), we show the mid-intensity level ($L=0.6$ for the first two, low-tertile for \Cat{Entity Swap}).

\subsection{Metric Configuration}
\label{appx:metric-config}

We describe each metric and the specific checkpoint or configuration used.

\paragraph{\chrf and \chrfpp~\citep{popovic-2015-chrf}.}
F-scores over character n-grams, with optional word n-grams in \chrfpp. They are less sensitive to tokenization and morphology than BLEU and perform well for morphologically rich languages, though they remain dependent on reference surface form rather than semantics. We use SacreBLEU's default tokenizers, \texttt{13a} for German, \texttt{zh} for Chinese, and MeCab-Ko for Korean.

\paragraph{\bertscore~\citep{zhang-2020-bertscore}.}
Cosine similarity between contextualized embeddings of candidate and reference, capturing both lexical and semantic similarity. We use \texttt{xlm-roberta-large} as the underlying model.

\paragraph{\unite and \uniteqe~\citep{wan-etal-2022-unite}.}
A unified framework that supports both QE and reference-based evaluation via contrastive training, designed to reduce overreliance on the reference while still benefiting from it when available. We use the \texttt{Unbabel/unite-mup} checkpoint.

\paragraph{\xcomet~\citep{guerreiro-etal-2024-xcomet}.}
A multilingual COMET variant built on Tower-based models, fine-tuned for multilingual MT evaluation. We use \texttt{Unbabel/XCOMET-XXL}.

\paragraph{\metricx~\citep{juraska-etal-2024-metricx}.}
A regression-based metric built on mT5 that predicts MQM-style quality scores. We use the \texttt{metricx-24-hybrid-xxl} checkpoint.

\paragraph{\xcometkiwi~\citep{rei-etal-2023-scaling} and \metricxqe~\citep{juraska-etal-2024-metricx}.}
Reference-free counterparts of \xcomet and \metricx that score translations from the source alone. They serve as a robustness reference point that is unaffected by reference perturbation.

%%%%%%%%%%%%%%%%%%%%%%%%%%%%%%%%%%%%%%%%%%%%%%%%%

\begin{table*}[t]
\centering
\small
%\resizebox{0.7\linewidth}{!}{
\begin{tabular}{l r r r r r r r r}
\toprule
 & \multicolumn{4}{c}{$\Delta_\textsc{ter}$} & \multicolumn{4}{c}{$d_\text{\ter}$} \\
\cmidrule(lr){2-5} \cmidrule(lr){6-9}
\textbf{Contamination} & \textbf{Zh} & \textbf{De} & \textbf{Ko} & \textbf{Avg} & \textbf{Zh} & \textbf{De} & \textbf{Ko} & \textbf{Avg} \\
\midrule
\WrongLangCat  & 16.5 & 24.0 & 35.1 & 25.2 & 23.9 & 27.3 & 47.0 & 32.7 \\
\ContentAddCat & 17.1 & 23.3 & 32.5 & 24.3 & 26.1 & 27.2 & 46.9 & 33.4 \\
\UnalignedCat  & 16.5 & 22.4 & 33.2 & 24.0 & 23.3 & 25.3 & 45.2 & 31.3 \\
\PolarityCat   & 16.2 & 22.7 & 31.9 & 23.6 & 25.1 & 26.2 & 44.6 & 32.0 \\
\MinorNoiseCat & 15.5 & 21.8 & 32.5 & 23.3 & 22.7 & 25.3 & 44.9 & 31.0 \\
\SourceCopyCat & 15.6 & 22.3 & 31.3 & 23.1 & 22.9 & 25.3 & 43.2 & 30.5 \\
\rowcolor{yellow!40}
\EntitySwapCat & -0.8 &  1.2 &  3.9 &  1.4 &  6.9 &  5.3 & 11.5 &  7.9 \\
\bottomrule
\end{tabular}
%}
\caption{Model behavior under contaminated references for \mQwen. Larger values
indicate stronger adherence to the reference at generation time.
$\Delta_\ter$ measures how much further the contaminated-reference
translation drifts from the clean reference compared to the clean-reference
baseline, and $d_\ter$ is the direct distance between the two outputs.
Rows are sorted by average $\Delta_\ter$.}
\label{tab:model_ter}
\end{table*}

\section{Further Study: Generation Under a Contaminated Reference}
\label{sec:model_behavior}

\paragraph{What this experiment is for.}
The metric results establish that reference contamination shifts metric scores.
They do not say whether the contaminated content is salient to a translation
system in the same way. We test that by giving a translation model the same
contaminated references we give the metrics and measuring how far its output
moves. Holding the input identical across the two is what lets a difference in
response be attributed to what each attends to rather than to what each was
shown.
 
Supplying a reference at inference time is not how translation systems are
deployed, and we do not present this as a realistic setting. Post-editing, where
a model revises an existing translation rather than being conditioned on a gold
reference, poses the same question in a deployed form, and we leave it to future
work.

\paragraph{Setup.}
We run \mQwen \citep{qwen3.5} on the full \mirrorbench across all three language pairs, prompting the model to translate each English source with either the clean or a contaminated reference shown in the prompt. Prompts are listed in \Cref{appx:prompts}.

\paragraph{Measures.}
We compute two complementary \ter-based measures~\citep{snover-etal-2006-study}. \enlist{i} The reference gap $\Delta_\text{\ter}$ compares the contaminated-reference translation to the clean reference, taking the clean-reference output as the baseline. Given a translation $T$ and a reference $R$,
\begin{equation*}
\Delta_\text{\ter} = \text{\ter}(T_\text{bias}, R_\text{clean}) - \textsc{ter}(T_\text{clean}, R_\text{clean}).
\end{equation*}
This isolates the effect of contamination on the output's distance from the clean reference. \enlist{ii} The output distance $d_\text{\ter}$ measures the edit distance between the two outputs generated under different reference conditions,
\begin{equation*}
d_\text{\ter} = \text{\ter}(T_\text{bias}, T_\text{clean}).
\end{equation*}
This captures how sensitively the model's output changes when only the reference changes. Larger values of both measures indicate stronger reference adherence at generation time. Both are computed with SacreBLEU ~\citep{post-2018-call} using \verb|asian_support=True, normalized=True|.

\paragraph{The model follows the contamination, except for \protect\Cat{Entity Swap}.}
\Cref{tab:model_ter} gives both measures. Six of the seven categories fall in a
narrow band, with average $\Delta_\textsc{ter}$ between $23.1$ and $25.2$ and no
category separating from the others by more than two points. \Cat{Entity Swap}
sits at $1.4$, an order of magnitude below, and its $d_\textsc{ter}$ drops from an
average of $30$ to $34$ down to $7.9$. The gap holds in all three language pairs
and is largest in Chinese, where $\Delta_\textsc{ter}$ turns slightly negative at
$-0.8$. The model behaves as though an entity-level contradiction with the source
is enough to discount that side of the input regardless of how large the
contradiction is. \ter alone does not identify which signal the model falls back
on, but the low $d_\textsc{ter}$ points to the source. This near-invariance holds
for the same contamination that most penalizes the reference-based metrics.

\paragraph{Stronger language priors resist reference bias.}
The three language pairs order consistently on both measures, with
En$\rightarrow$Zh lowest and En$\rightarrow$Ko highest across every category in
\Cref{tab:model_ter}. \mQwen is Chinese-heavy in pre-training, which we take as a
possible reason its Chinese generation is more source-anchored. Testing that would
require controlled training data. The ordering is consistent with a translation
model trained more heavily on the target language resisting reference bias at
generation time, and we report it as a pattern rather than as a result.
\subsection{Prompts}
\label{appx:prompts}

We list the prompts used for LLM-based contamination generation. Both \Cat{Polarity Flip} and \Cat{Entity Swap} use \mGpt with \texttt{reasoning\_effort=low} via the OpenAI Batch API. 

\Cat{Polarity Flip} reverses the polarity of a target-language reference at the sentence level. The model is instructed to flip affirmative statements to negative and vice versa while preserving fluency, length, and lexical choice.

\Cat{Entity Swap} first extracts named entities with spaCy, then prompts \mGpt to generate contextually plausible replacements of the same entity type. The prompt is zero-shot with the target language name (Korean, German, or Chinese) inserted via a placeholder. When the same entity appears multiple times in a segment, each occurrence receives a distinct replacement so that entity coreference is disrupted.

\newpage
\begin{promptbox}{Polarity Flip Prompt}
You are a linguistic annotator. Reverse the polarity of the following sentence in \{tgt\_lang\}.

\vspace{0.4em}
Rules:
\begin{itemize}[leftmargin=*, itemsep=0pt, topsep=0pt]
\item Flip affirmative to negative, and negative to affirmative.
\item Keep the original sentence structure, word order, and length as close to the source as possible.
\item Do not change any named entities, numbers, or content words other than what is required for polarity.
\item Output only the reversed sentence with no explanation.
\end{itemize}

\vspace{0.4em}
Input: \{sentence\}

\vspace{0.4em}
Output:
\end{promptbox}

\begin{promptbox}{Entity Swap Prompt}
\textbf{System:} You are a multilingual cultural and linguistic expert.

\textbf{User:}

You are a \{lang\} cultural and linguistic expert.

Given a list of named entity occurrences from a \{lang\} text, suggest a DIFFERENT entity of the SAME TYPE for each occurrence.

Rules:
\begin{itemize}[leftmargin=*, itemsep=0pt, topsep=0pt]
\item Each occurrence must receive a UNIQUE replacement, even if the original text is identical across occurrences. Two occurrences of the same person name must each get a different person name.
\item Keep the replacement in the SAME LANGUAGE and SCRIPT as the original.
\item PER (person): a different person name natural in the \{lang\}-speaking world.
\item LOC (location): a different place name appropriate to the context.
\item ORG (organization): a different organization name.
\item Return ONLY a JSON object mapping occurrence index (1-based) to its replacement.
\end{itemize}

Entities:

1. "\{entity\_1\}" (\{type\_1\})

2. "\{entity\_2\}" (\{type\_2\})

...

Return format:

\{"1": "<replacement\_1>", "2": "<replacement\_2>", ...\}
\end{promptbox}

\begin{promptbox}{Translation with Reference Prompt}
Translate the following English sentence into \{tgt\_lang\}. A reference translation is provided to guide you, but you should rely on the source sentence as the primary signal.

\vspace{0.4em}
Source: \{src\_seg\}

Reference: \{ref\_seg\}

\vspace{0.4em}
Return ONLY a JSON object with a single field 'translation' containing the \{tgt\_lang\} translation. Do not include any other keys, explanation, or formatting.

\vspace{0.4em}
Format: \{"translation": "<your \{tgt\_lang\} translation>"\}
\end{promptbox}
\begin{table*}[t]
\centering
\small
\setlength{\tabcolsep}{4pt}
\renewcommand{\arraystretch}{1.15}
\resizebox{\linewidth}{!}{
\begin{tabular}{l c c c c c c c}
\toprule
 & \MinorNoiseCat & \SourceCopyCat & \WrongLangCat & \UnalignedCat & \ContentAddCat & \PolarityCat & \EntitySwapCat \\
 & VR  /  $\Delta$ & VR  /  $\Delta$ & VR  /  $\Delta$ & VR  /  $\Delta$ & VR  /  $\Delta$ & VR  /  $\Delta$ & VR  /  $\Delta$ \\
\midrule
\multicolumn{8}{l}{\textit{Reference-based}} \\
\chrf & .00 / +24.1 & .00 / +25.2 & .00 / +52.6 & .00 / +30.4 & .00 / +14.7 & .06 / +5.09 & \sMax{.98 / -32.8} \\
\bertscore & .00 / +.116 & .00 / +.094 & .00 / +.065 & .04 / +.072 & .00 / +.022 & .06 / +.011 & \sMax{.94 / -.034} \\
\unite & .00 / +1.04 & .00 / +1.08 & .03 / +.586 & .06 / +1.01 & .02 / +.585 & .01 / +.447 & \sHi{.56 / +.010} \\
\xcomet & .01 / +.362 & .00 / +.427 & \sLow{.10 / +.153} & .03 / +.461 & .02 / +.086 & .07 / +.048 & \sLow{.19 / +.059} \\
\metricx & .00 / +8.62 & .00 / +14.5 & \sLow{.17 / +.673} & .03 / +9.07 & .00 / +7.96 & .04 / +1.33 & \sLow{.12 / +1.83} \\
\bottomrule
\end{tabular}
}
\caption{Per-language robustness scores for \textsc{En-De}.}
\label{tab:lang-en-de}
\end{table*}

\begin{table*}[t]
\centering
\small
\setlength{\tabcolsep}{4pt}
\renewcommand{\arraystretch}{1.15}
\resizebox{\linewidth}{!}{
\begin{tabular}{l c c c c c c c}
\toprule
 & \MinorNoiseCat & \SourceCopyCat & \WrongLangCat & \UnalignedCat & \ContentAddCat & \PolarityCat & \EntitySwapCat \\
 & VR  /  $\Delta$ & VR  /  $\Delta$ & VR  /  $\Delta$ & VR  /  $\Delta$ & VR  /  $\Delta$ & VR  /  $\Delta$ & VR  /  $\Delta$ \\
\midrule
\multicolumn{8}{l}{\textit{Reference-based}} \\
\chrf & .00 / +12.7 & .00 / +20.8 & .01 / +30.1 & .00 / +22.1 & .00 / +9.28 & .07 / +4.76 & \sMax{1.00 / -52.9} \\
\bertscore & .01 / +.064 & .00 / +.078 & .00 / +.055 & .03 / +.061 & .00 / +.020 & .03 / +.011 & \sMax{.97 / -.042} \\
\unite & .03 / +.472 & .00 / +.824 & .03 / +.457 & .06 / +.911 & .02 / +.503 & .02 / +.377 & \sHi{.67 / -.132} \\
\xcomet & .03 / +.197 & .01 / +.358 & \sMid{.21 / +.120} & .05 / +.486 & .03 / +.144 & .06 / +.095 & \sMid{.26 / +.080} \\
\metricx & .02 / +4.99 & .00 / +10.2 & \sLow{.15 / +.757} & .04 / +8.68 & .00 / +7.82 & .03 / +1.35 & \sMid{.27 / +1.08} \\
\bottomrule
\end{tabular}
}
\caption{Per-language robustness scores for \textsc{En-Ko}.}
\label{tab:lang-en-ko}
\end{table*}

\begin{table*}[t]
\centering
\small
\setlength{\tabcolsep}{4pt}
\resizebox{\linewidth}{!}{
\renewcommand{\arraystretch}{1.15}
\begin{tabular}{l c c c c c c c}
\toprule
 & \MinorNoiseCat & \SourceCopyCat & \WrongLangCat & \UnalignedCat & \ContentAddCat & \PolarityCat & \EntitySwapCat \\
 & VR  /  $\Delta$ & VR  /  $\Delta$ & VR  /  $\Delta$ & VR  /  $\Delta$ & VR  /  $\Delta$ & VR  /  $\Delta$ & VR  /  $\Delta$ \\
\midrule
\multicolumn{8}{l}{\textit{Reference-based}} \\
\chrf & .00 / +22.4 & .02 / +30.7 & .00 / +37.4 & .00 / +28.3 & .00 / +15.0 & .05 / +6.14 & \sMax{.97 / -44.1} \\
\bertscore & .00 / +.101 & .01 / +.127 & .00 / +.061 & .04 / +.067 & .00 / +.028 & .06 / +.010 & \sMax{.96 / -.036} \\
\unite & .00 / +.717 & .01 / +.971 & .06 / +.427 & .07 / +.876 & .04 / +.613 & .02 / +.362 & \sMax{.71 / -.109} \\
\xcomet & .02 / +.417 & .03 / +.320 & .04 / +.256 & .04 / +.442 & .05 / +.138 & .09 / +.075 & \sMid{.35 / +.041} \\
\metricx & .00 / +10.7 & .02 / +3.99 & .10 / +.921 & .06 / +3.59 & .01 / +5.90 & .06 / +.790 & \sMid{.30 / +.629} \\
\bottomrule
\end{tabular}
}
\caption{Per-language robustness scores for \textsc{En-Zh}.}
\label{tab:lang-en-zh}
\end{table*}
\end{document}